\documentclass[sn-mathphys-ay]{sn-jnl}

\usepackage{graphicx}%
\usepackage{multirow}%
\usepackage{amsmath,amssymb,amsfonts}%
\usepackage{amsthm}%
\usepackage{mathrsfs}%
\usepackage[title]{appendix}%
\usepackage{xcolor}%
\usepackage{textcomp}%
\usepackage{manyfoot}%
\usepackage{booktabs}%
\usepackage{url}
\usepackage{listings}%
\usepackage{multirow}
\usepackage[ruled,vlined,linesnumbered]{algorithm2e}
\usepackage{optidef}
\usepackage{hyperref}
\usepackage{placeins}
\usepackage{url}
\usepackage{cleveref}
\usepackage{tabularx} 
\usepackage{subcaption}

\begin{document}

\title[Random-Key Optimizer]{A Random-Key Optimizer for\linebreak Combinatorial Optimization}

\author*[1]{\fnm{Antonio A.} \sur{Chaves}}\email{antonio.chaves@unifesp.br}

\author[1,3]{\fnm{Mauricio G.C.} \sur{Resende}}\email{mgcr@berkeley.edu}

\author[2]{\fnm{Martin J.A.} \sur{Schuetz}}\email{maschuet@amazon.com}

\author[2]{\fnm{J. Kyle} \sur{Brubaker}}\email{johbruba@amazon.com}

\author[2]{\fnm{Helmut G.} \sur{Katzgraber}}\email{katzgrab@amazon.com}

\author[4]{\fnm{Edilson F. de} \sur{Arruda}}\email{e.f.arruda@southampton.ac.uk}

\author[5]{\fnm{Ricardo M. A.} \sur{Silva}}\email{rmas@cin.ufpe.br}

\affil*[1]{\orgname{Federal U. of São Paulo}, \orgaddress{\city{S. J. dos Campos},  \state{SP}, \country{Brazil}}}

\affil[2]{\orgdiv{Amazon Advanced Solutions Lab}, \orgaddress{\city{Seattle},  \state{WA}, \country{USA}}}

\affil[3]{\orgname{DIMACS}, \orgaddress{\city{Piscataway},  \state{NJ}, \country{USA}}}

\affil[4]{\orgname{University of Southampton}, \orgaddress{\city{Southampton},  \state{Hampshire}, \country{UK}}}

\affil[5]{\orgname{Federal U. of Pernambuco}, \orgaddress{\city{Recife}, \state{PE}, \country{Brazil}}}


\abstract{
This paper introduces the Random-Key Optimizer (RKO), a versatile and efficient stochastic local search method tailored to combinatorial optimization problems. Using the random-key concept, RKO encodes solutions as vectors of random keys that are subsequently decoded into feasible solutions via problem-specific decoders. The RKO framework is able to combine a plethora of classic metaheuristics, each capable of operating independently or in parallel, with solution sharing facilitated through an elite solution pool. This modular approach allows for the adaptation of various metaheuristics, including simulated annealing, iterated local search, and greedy randomized adaptive search procedures, among others. The efficacy of the RKO framework, implemented in C++ and publicly available\footnote{Github public repository: \url{github.com/RKO-solver}}, is demonstrated through its application to three NP-hard combinatorial optimization problems: the $\alpha$-neighborhood $p$-median problem, the tree of hubs location problem, and the node-capacitated graph partitioning problem. The results highlight the framework's ability to produce high-quality solutions across diverse problem domains, underscoring its potential as a robust tool for combinatorial optimization.

}

\maketitle

\setcounter{algocf}{0}  

\section{Introduction}\label{sec:intro}

An \textit{instance}
of a \textit{combinatorial optimization problem} is
defined by a 
finite \textit{ground set}
$E=\{1,\dots,n\}$, a set of
{feasible solutions}
$F \subseteq 2^E$, and an
{objective function}
$f: 2^{E} \rightarrow \mathbb{R}$.
In the case of a minimization (maximization) problem, we seek a \textit{global optimal solution} $S^* \in F$ such that $f(S^*) \leq f(S)$ ($f(S^*) \geq f(S)$), \; $\forall S \in F$.
The ground\index{ground set}
set $E$, the cost function $f$, and the set of feasible solutions
$F$ are defined for each specific problem.

In the traveling salesman problem (TSP) on a graph $G=(N,A)$, for example, one seeks the shortest tour of arcs in $A$ that visits each node in $N$ exactly once and returns to the first node. The ground set $E$ for the TSP consists of the sets of $|A|$ arcs
while the set of feasible solutions is made up of all subsets of arcs in $A$ such that they form a tour of the nodes in $N$.  The cost $f$ of a tour is the sum of the lengths of the arcs in the tour.

A \textit{random key} $x$ is a real number in the interval $[0,1)$, i.e. $x \in [0,1) \subseteq \mathbb{R}$. A vector $\chi$ of $n$ random keys is a point in the unit hypercube in $\mathbb{R}^n$, $\chi=(x_1,x_2, \ldots, x_n)$, where $x_i \in [0,1)$.
We shall refer to a vector of $n$ random keys simply as \textit{random keys}. 
A solution of a combinatorial optimization problem can be encoded with random keys. Such random key vectors may be seen as an abstract representation of the solution in some latent (embedding) space, akin to \textit{embedding} vectors, thus providing standardized representations of data designed to be consumed by machine learning models \citep{bishop2023deep}. Given a vector $\chi$
of random keys, a \textit{decoder} $\mathcal{D}$ takes as input $\chi$ and outputs a feasible solution
$S \in F$ of the combinatorial optimization problem, i.e., $F = \mathcal{D}(\chi)$.

The Random-Key Optimizer (RKO) is a stochastic local search method that employs the random-key concept for solution representation to address combinatorial optimization problems. Since the introduction of the first random-key genetic algorithm by \cite{Bean_RKGA_1994}, followed by the biased random-key genetic algorithm (BRKGA) of \citet{Gonçalves_BRKGA_2011}, various metaheuristics have been adapted to this framework. 
Recent adaptations include 
particle swarm optimization \citep{bewoor2018production},
harmony search \citep{garcia2015random},
iterated local search and iterated greedy search \citep{pessoa2018heuristics},
path-relinking \citep{andrade2021multi},
dual annealing \citep{Schuetz_RKO_2022}, 
variable neighborhood search \citep{ManPolMacJulProGiaSalCha23a}, and the greedy randomized adaptive search procedure \citep{RK-GRASP}.

This paper presents a C++ implementation of the RKO framework, which simplifies user interaction by requiring only the development of a decoder function. The current framework incorporates eight classic metaheuristics that can operate independently or in parallel, with the latter approach facilitating solution sharing through an elite solution pool. These metaheuristics are problem-independent, relying on the decoder to map the random-key space to the solution space of the specific optimization problem. Additional metaheuristics can be easily added to the framework. As a proof of concept, RKO was tested on three NP-hard combinatorial optimization problems: the $\alpha$-neighborhood $p$-median problem, the tree of hubs location problem, and the node-capacitated graph partitioning problem.

The structure of this paper is as follows. To illustrate the idea of encoding and decoding with random keys, \Cref{sec:examples} first introduces decoders from successful applications.
\Cref{sec:rko} introduces the Random-Key Optimizer (RKO) concept. \Cref{sec:framework} details the RKO framework components, including metaheuristics, shaking, blending, and local search modules. \Cref{sec:aplications} demonstrates the application of RKO to three distinct combinatorial optimization problems, each utilizing a different decoder.
Finally, \Cref{sec:conclusion} offers concluding remarks.

\section{Encoding and decoding with random keys}
\label{sec:examples}

The random-key representation confers significant advantages in solving complex combinatorial optimization problems when coupled with problem-dependent decoders. This approach preserves solution feasibility, simplifies search operators, and enables the development of problem-independent metaheuristics. This paradigm facilitates efficient navigation of highly constrained solution spaces by establishing a mapping between continuous and discrete domains. Furthermore, it stimulates the creation of adaptable optimization algorithms applicable across diverse optimization problems, allowing for core search mechanisms while adapting problem-specific constraints through customized decoders.

In the next subsections, we illustrate the encoding and decoding processes for a diverse range of application domains, including packing, vehicle routing, and internet traffic engineering.

\subsection{Traveling Salesman Problem}

\citet{Bean_RKGA_1994} first proposed random key encoding for problems whose solutions can be represented as a permutation vector, as is the case for the TSP, an NP-hard problem \citep{Karp1972}. Given a vector of random keys $\chi$, the decoder simply sorts the keys of the vector, and the indices of the sorted vector represent a permutation of ${1,2,\ldots,n}$.

\begin{figure}[ht]
    \centering
    \includegraphics[width=0.28\linewidth]{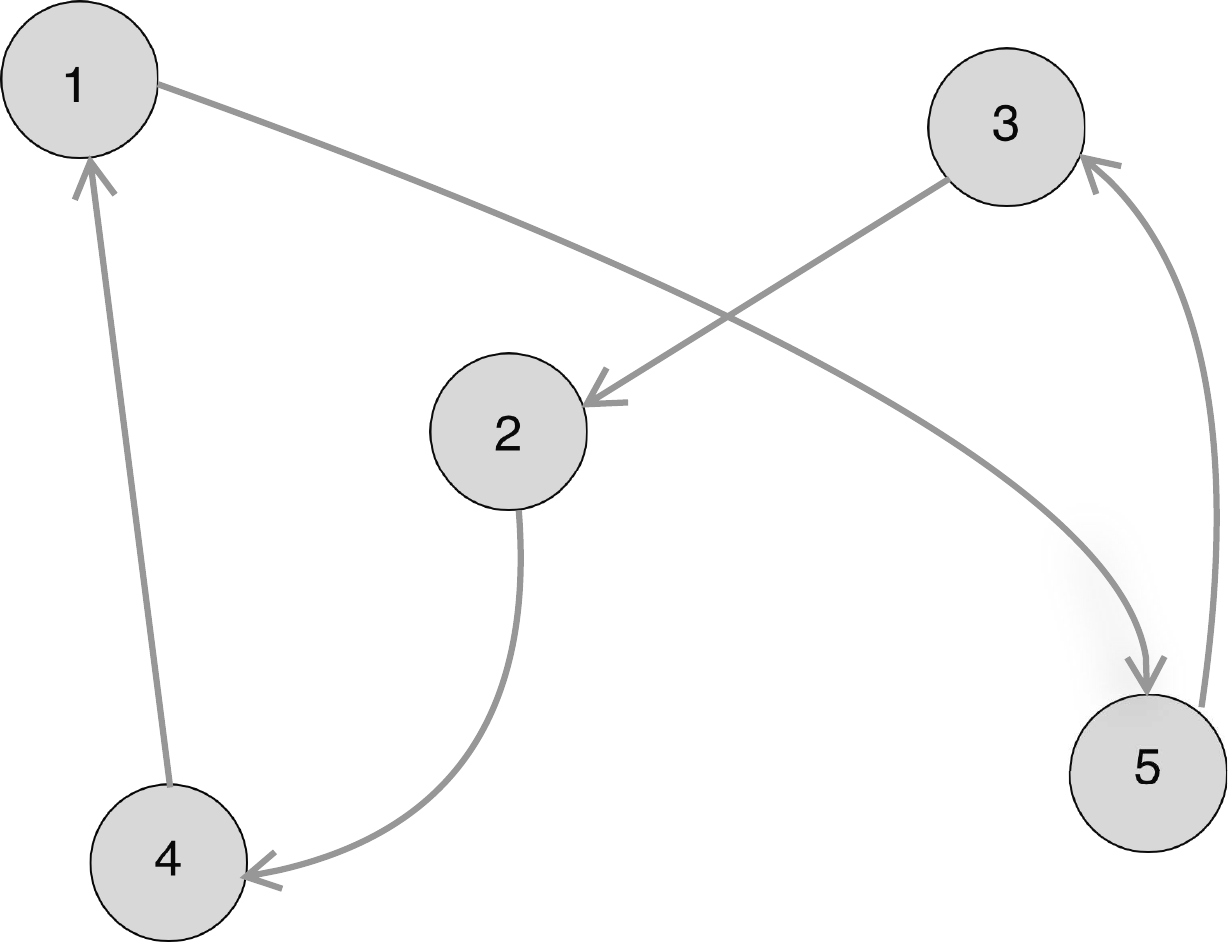}
    \caption{A TSP tour, decoded from the vector of random keys $\chi = (0.085, 0.277, 0.149, 0.332, 0.148)$.}
    \label{fig:TSP-tour}
\end{figure}

Consider a random key vector $\chi = (0.085, 0.277, 0.149, 0.332, 0.148)$.
Sorting the vector in increasing order, we get $\sigma[\chi]= (0.085, 0.148, 0.149, 0.277, 0.332)$ with corresponding indices
$\pi(\sigma[\chi]) = (1, 5, 3, 2, 4)$.  \Cref{fig:TSP-tour} shows this tour where we start at node 1, then visit nodes 5, 3, 2, and 4, in this order, and finally return to node 1.

\subsection{Set Covering Problem}
\label{s_covering}

Given $n$ finite sets $P_1, P_2, \ldots, P_n$, let
sets $I$ and $J$ be
defined as
$I = \displaystyle{\cup_{j=1}^n P_j} = \{1,2,\ldots,m\}$ and
$J = \{1,\ldots,n\}$.
A subset 
$J^* \subseteq J$
is called a \textit{cover} if $\displaystyle \cup_{j \in J^*} P_j = I$.
The \textit{set covering problem} is to find a 
cover of minimum cardinality.
Let $A$ be the binary $m \times n$ matrix such that
$A_{i,j} = 1$ if and only if $i \in P_j$.
An integer programming formulation for set covering is
\begin{displaymath}
\min \; \{e_n x \;:\; A x \geq e_m,\; x \in \{0,1\}^n\},
\end{displaymath}
where $e_k$ denotes a vector of $k$ ones and $x$ is a
binary $n$-vector such that $x_j = 1$ if and only if
$j \in J^*$.
The set covering problem has many applications, such as crew scheduling, cutting stock, facilities location, and others \citep{Vem98a} and
is NP-hard \citep{Karp1972,GarJoh79a}.

Random keys can be used to encode solutions of the set covering problem. In one approach, the vector of $n$ random keys $\chi$ is sorted, and elements are added to the cover in the order given by $\pi(\sigma(\chi))$ until a cover is constructed. Then, elements are scanned in the same order, and each element is tentatively removed from the cover. The removal is made permanent if the cover is not destroyed by its removal.

Another approach \citep{ResTosGonSil12a} makes use of a greedy algorithm. As with the first decoder, this decoder takes the vector of random keys $\chi$ as input and returns a cover $J^* \subseteq J$. To describe the decoding procedure, let the cover be represented by a binary vector $\mathcal{Y} = (\mathcal{Y}_1, \ldots,\mathcal{Y}_{|J|})$, where $\mathcal{Y}_j = 1$ if and only if $j \in J^*$.

The decoder has three phases.  
In the first phase, for $j=1,\ldots,|J|$, the values of $\mathcal{Y}_j$ are initially
set according to 
\begin{displaymath}
\mathcal{Y}_j = 
\begin{cases}
1 & \text{if $\chi_j \geq 0.5$}\\
0 & \text{otherwise.}
\end{cases}
\end{displaymath}
The indices implied by the binary vector $\mathcal{Y}$ can correspond
to either a feasible or infeasible cover $J^*$.
If $J^*$ is a feasible cover, then the second phase is skipped.
If $J^*$ is not a valid cover, then the second phase of the decoding procedure builds a valid cover with
the greedy algorithm for set covering of \citet{Joh74a},
starting from the partial cover $J^*$ defined by $\mathcal{Y}$.
This greedy algorithm proceeds as follows.
While $J^*$ is not a valid cover, select
the smallest index $j \in J \setminus J^*$
for which the inclusion of $j$ in $J^*$
covers a maximum number of yet-uncovered elements of $I$.  
The third phase of the decoder attempts to remove superfluous elements from cover $J^*$ as in the case of the first decoder described above.
While there is some element $j \in J^*$ such that $J^* \setminus \{j\}$
is still a valid cover, then such an element having the smallest index is removed from $J^*$. 
Figure \ref{fig:SetCovering} shows three phases of the set covering decoding with $n=9$ and $m=5$.

\begin{figure}[ht]
    \centering
    \includegraphics[width=\linewidth]{CoveringExample.pdf}
    \caption{Example of decoding a set covering problem. $A$ is an input matrix for a set covering problem whose solution vector $X$ of selected sets is reconstructed from an embedded vector $Y$.}
    \label{fig:SetCovering}
\end{figure}

\subsection{OSPF routing in intradomain traffic engineering}

Let $G=(N,A)$ be an Internet Protocol (IP) network, where $N$ is its set of router nodes and $A$ is its set of links.
Given a set of traffic demands between origin-destination (O-D) pairs in the network, the Open Shortest Path First (OSPF), an NP-hard problem \citep{giroire2013hardness}, the weight setting problem consists in determining weights to be assigned to the links so as to optimize a cost function, typically associated with a network congestion measure, when traffic is routed on least-weight paths between O-D pairs. Link weights $w_1, w_2, \ldots, w_{|A|}$ typically are
integer-valued in the interval $[1, \bar{w}]$, where $\bar{w} = 2^{16}-1$.

Solutions to the OSPF weight setting problem are encoded with a vector $\chi$ of $|A|$ random keys \citep{EriResPar02a,BurResRibTho05a}.
The decoder first sets link weights as $w_i = \lceil \chi_i \times \bar{w} \rceil$, for $i=1,2,\ldots,|A|$, where $\lceil z \rceil | z \in \mathbb{R}$ is the ceiling function that returns the smallest integer greater than or equal to $z$.
Then, the demand for each O-D pair is routed on a least-weight path.
For each link $a \in A$, the flows from each O-D pair on that link are summed up, resulting in the total flow $F_a$ on link $a$.
The link congestion cost $\phi_a(F_a)$ is computed for each link $a\in A$ and the total network
congestion cost is then computed as $\Phi = \sum_{a \in A}\phi_a(F_a)$.

\subsection{Redundant content distribution}

\citet{JohEtAlii2020} consider a situation where we need to distribute data, like video-on-demand, over a network where link or vertex failures cannot be quickly repaired. These failures could cause costly service interruptions, so we want a robust distribution process that is resilient to any single vertex or link failure. To achieve this, we need to place multiple copies of our data source in the network. However, due to hosting costs, we want to minimize the number of hosting sites rather than placing a copy at each service hub.

The \textit{setwise-disjoint facility location} model applies when we do not control routing, relying on the network's shortest-path protocols (like OSPF) instead. Here, to guarantee vertex-disjoint paths to a customer from two facility locations, we must ensure that their shortest path sets intersect only at the customer location. We can examine every pair of facility locations, $s$ and $t$, and every customer location $u$.  If the shortest paths from $s$ to $u$ and the shortest paths from $t$ to $u$ only intersect at $u$, then $s$ and $t$ cover $u$ and the triple $(u,s,t)$ can be saved for possible use in a solution to the setwise-disjoint facility location problem.  Model the network as a graph $G=(N,A)$, where $N$ are the vertices of $G$ and $A$ are its links.
Let $S \subseteq N$ be the set of nodes where hosting facilities can be located, and assume users are located on any node belonging to the set $U \subseteq N$.

In the \textit{set cover by pairs problem}, we are
given a ground set $U$ of elements, a set $S$ of {\em cover objects},
and a set $T$ of triples $(u,s,t)$, where $u \in U$ and $s,t \in S$.
We seek a minimum-cardinality cover by pairs subset $S^* \subseteq S$ for $U$, where $S^*$
{\em covers} $U$ if for each $u \in U$, there are $s,t \in S^*$ such
that $(u,s,t) \in T$.

Solutions to the setwise-disjoint facility location problem can be encoded with a vector $\chi$ of $|S|$ random keys.
The decoding process is similar to the second decoder for set covering, as introduced in \Cref{s_covering}, with the key difference that it introduces pairs of elements at a time instead of one element at a time.
This decoder takes as input the vector of random keys $\chi$ and returns a cover by pairs $S^* \subseteq S$.
Let the cover by pairs be a binary vector $\mathcal{Y} = (\mathcal{Y}_1, \ldots,\mathcal{Y}_{|S|})$, where $\mathcal{Y}_j = 1$ if and only if index $j$ is included in the cover $S^*$. The initial values of $\mathcal{Y}_j$ for $j=1,\ldots,|S|$ are $\mathcal{Y}_j = 1$ if $\chi_j \geq 0.5$ and $\mathcal{Y}_j = 0$ otherwise.

The binary vector $\mathcal{Y}$ represents a partial cover $S^*$ of $U$. The decoder proceeds to the next step if this partial cover is already feasible. However, if $S^*$ is not a valid cover, the second step of the decoding process begins. Following the greedy method for set cover by pairs from \citet{JohEtAlii2020}, a valid cover is constructed as follows:

\begin{enumerate}
    \item While $S^*$ is not a feasible cover, select the smallest index $j \in S \setminus S^*$ whose inclusion in $S^*$ covers the greatest number of uncovered elements in $U$.
    \item If no such element exists, find the smallest indexed pair $(i, j) \in S \setminus S^*$ whose inclusion in $S^*$ covers the maximum number of uncovered elements in $U$.
    \item If such a pair also does not exist, the problem is infeasible.
\end{enumerate}

In the third step of the decoder, superfluous elements are removed from cover $S^*$. This is done iteratively: if single elements $j \in S^*$ can be removed while maintaining a valid cover $S^* \setminus \{j\}$, we assemble such elements into a subset, select the element with the smallest index from this subset, and then remove this element from the set $S^*$.

\subsection{2D orthogonal packing}

In the two-dimensional non-guillotine packing problem, rectangular sheets must be packed into a larger rectangular sheet to maximize value. The rectangular sheets cannot overlap or be rotated and must align with the larger sheet's edges. This problem, which is NP-hard \citep{GarJoh79a}, is significant both theoretically and practically, with applications in industries like textiles, glass, steel, wood, and paper, where large sheets of material are cut into smaller rectangular pieces.

Given a large rectangular sheet of dimension $H \times W$ and $N$ types of smaller rectangular sheets, of dimensions $h_1 \times w_1, h_2 \times w_2, \ldots, h_N \times w_N$. There are $Q_i$ sheets of type $i$, each having value $V_i$.  Let $R_i$ denote the number of sheets of type $i$ that are packed into the larger rectangular sheet.  We seek a packing that maximizes the total value $\sum_{i=1,N} R_i \times V_i$, where $R_i \leq Q_i$, for 
$i=1,2,\ldots,N$.

\citet{GonRes11b} present an encoder and decoder for 2D orthogonal packing. Let $N'$ be the total number of available smaller rectangular
sheets, i.e. $N' = \sum_{i=1}^{N} Q_i$. Solutions are encoded with a vector $\chi$ of  $2N'$ random keys. To decode a solution, we sort the first $N'$ keys of $\chi$ in increasing order. The indices of the sorted keys impose a sequence for placement of the $N'$ rectangles. The decoder uses the last $N'$ keys to determine which of two placement heuristics is used to place rectangle $i$: Left-Bottom (LB) or Bottom-Left (BL).
Placement heuristic Left-Bottom takes the small rectangle from the top-right corner and moves it as far left as possible and then as far bottom
as possible while placement heuristic Bottom-Left also takes the small rectangle from the top-right corner but moves it as far down as possible and then as far left as possible. If $\chi_{N'+i} > \frac{1}{2}$, we pack rectangle $i$ with Bottom-Left placement, else we use Left-Bottom placement.
If the rectangle cannot be packed with either placement heuristic, it is simply discarded and the decoder moves on to the next small rectangle.

\begin{figure}[htbp]
    \centering
    \begin{subfigure}{0.3\linewidth}
        \centering
        \includegraphics[width=1.0\linewidth]{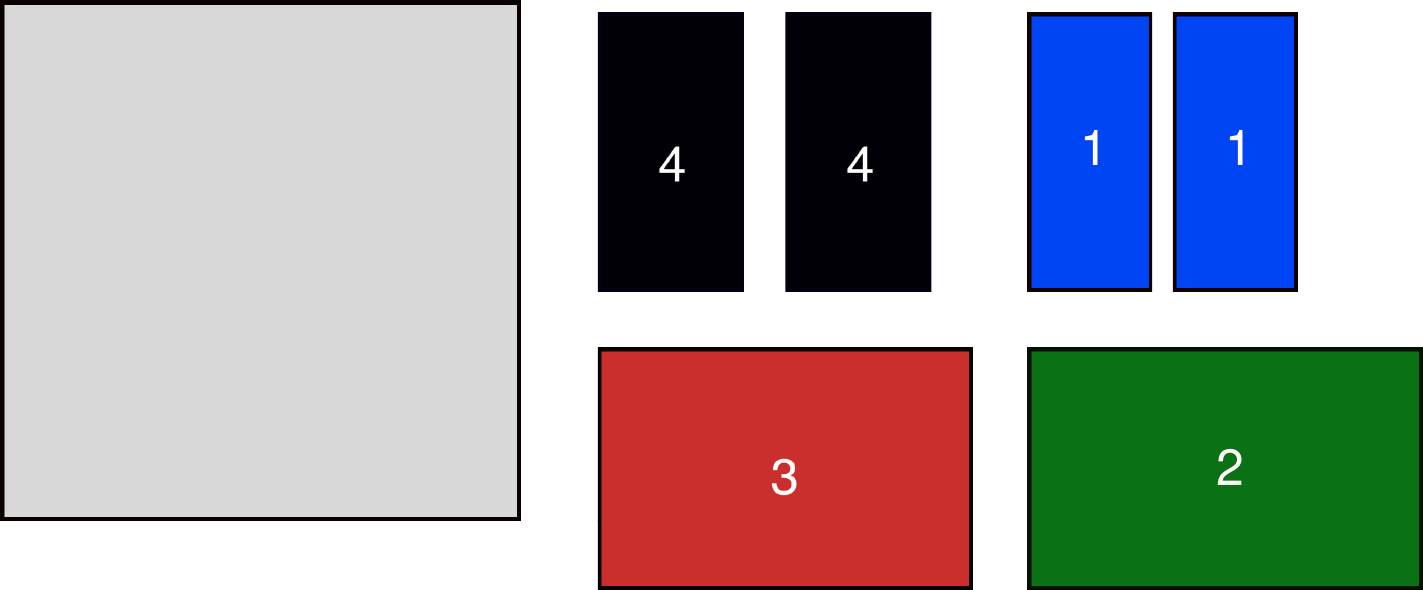}
        \caption{Four rectangle types are given, two of types 1 and 4, and one of types 2 and 3.}
        \label{fig:2DplacementINPUT}
    \end{subfigure}
    \hfill
    \begin{subfigure}{0.68\linewidth}
        \centering
        \includegraphics[width=1.0\linewidth]{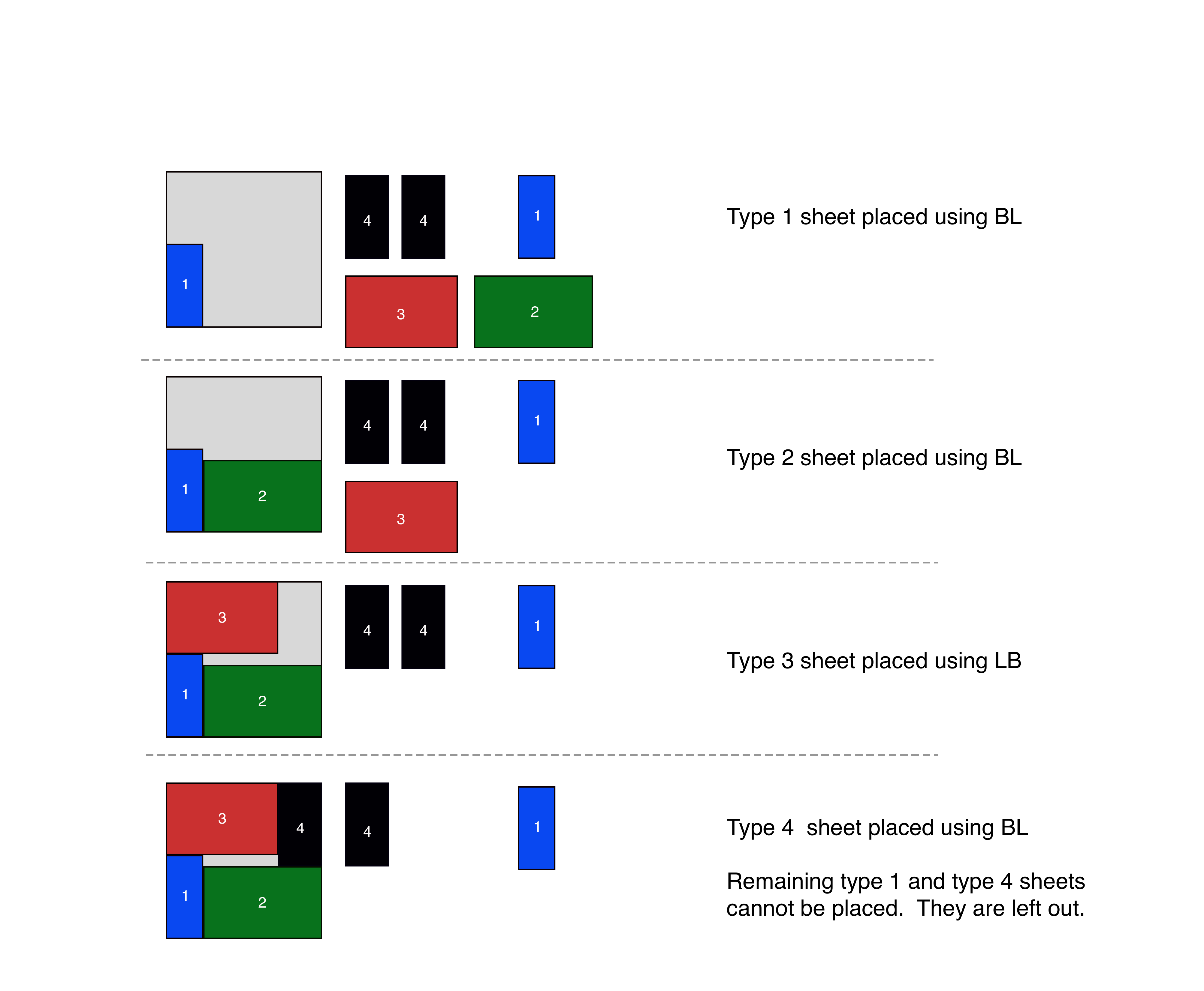}
        \caption{Decoding steps.}
        \label{fig:2D-layout-step-by-step}
    \end{subfigure}
    \caption{Example of input and decoder for 2D orthogonal packing.}
    \label{fig:2D}
\end{figure}

Figure~\ref{fig:2DplacementINPUT} shows an instance of the 2D orthogonal packing problem with four types of small rectangles where two of them have two copies while the other two are single copies. Therefore $N' = 6$.  Indices 1 and 2 correspond to the two type-1 rectangles.  Indices 3 and 4 correspond to, respectively, the type-2 and type-3 rectangles.  Finally, indices 5 and 6 correspond to the two type-4 rectangles.

Consider the encoded vector of random keys
\begin{displaymath}
    \chi = (.12, .90, .34, .55, .88, .99 \; | \; .63, .02, .98, .21, .99, .80).
\end{displaymath}
Let $\chi_1$ be the first $N' = 6$ keys of $\chi$. Sorting the keys of $\chi_1$, we get $\sigma(\chi_1) = (.12, .34, .55, .88, .90, .99)$ with
corresponding rectangle indices $\pi(\sigma(\chi_1)) = (1, 3, 4, 5, 2, 6)$ which gives us a placement order for the rectangles.
Using the last $N'$ keys we use placement heuristics BL, LB, BL, LB, BL, BL for, respectively, rectangles 1, 2, 3, 4, 5, and 6.

Figure~\ref{fig:2D-layout-step-by-step} shows four steps of the placement procedure.
On top, a type 1 rectangular sheet (rectangle 1) is placed using the BL heuristic.
Next, a type 2 rectangular sheet (rectangle 3) is placed using the BL heuristic.
Next, a type 3 rectangular sheet (rectangle 4) is placed using an LB heuristic.
Finally, on the bottom of the figure, a type 4 rectangular sheet (rectangle 5) is placed using a BL heuristic.
Neither of the two remaining rectangular sheets (one of type 1 and the other of type 4) can be added to the large sheet.
The total value of the solution is $\sum_{i=1,N} R_i \times V_i = V_1 + V_2 + V_3 + V_4.$

\subsection{Vehicle routing problem}

The vehicle routing problem (VRP) has many practical applications, e.g., in logistics. In these problems, one or more vehicles depart from a depot
and visit a number of customer nodes and then return to the depot.  One common objective is to minimize the cost of delivery, e.g., minimize total distance
given a fixed number of vehicles, or minimize the number of vehicles, given a maximum distance traveled by each vehicle.  Often, constraints are imposed
on the capacity of vehicles and on when they visit the customers.

\begin{figure}[htbp]
    \centering
    \begin{subfigure}{0.65\linewidth}
        \centering
        \includegraphics[width=1.0\linewidth]{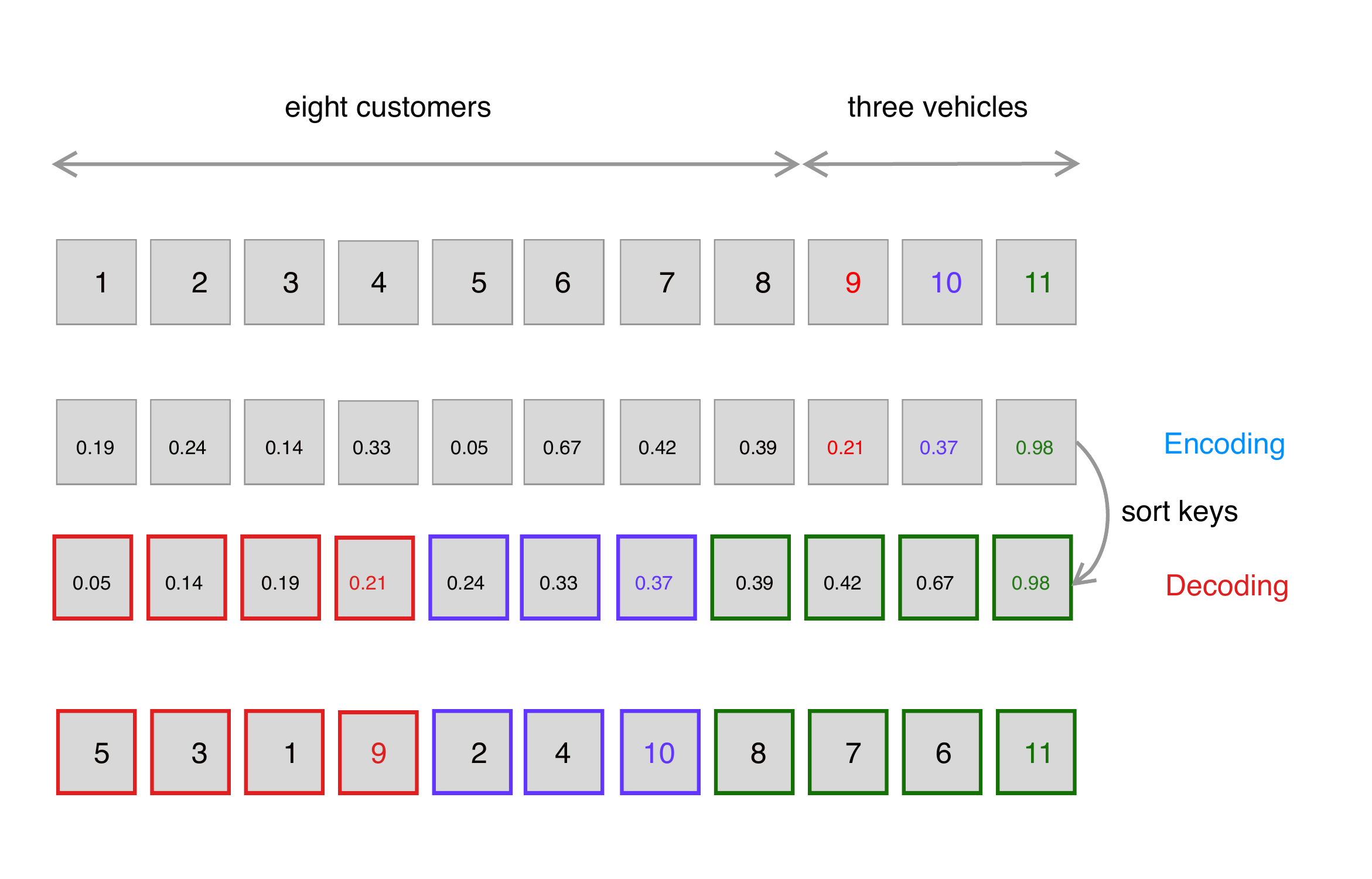}
        \caption{Decoding a VRP.}
        \label{fig:VRP-decoder}
    \end{subfigure}
    \hfill
    \begin{subfigure}{0.33\linewidth}
        \centering
        \includegraphics[width=1.0\linewidth]{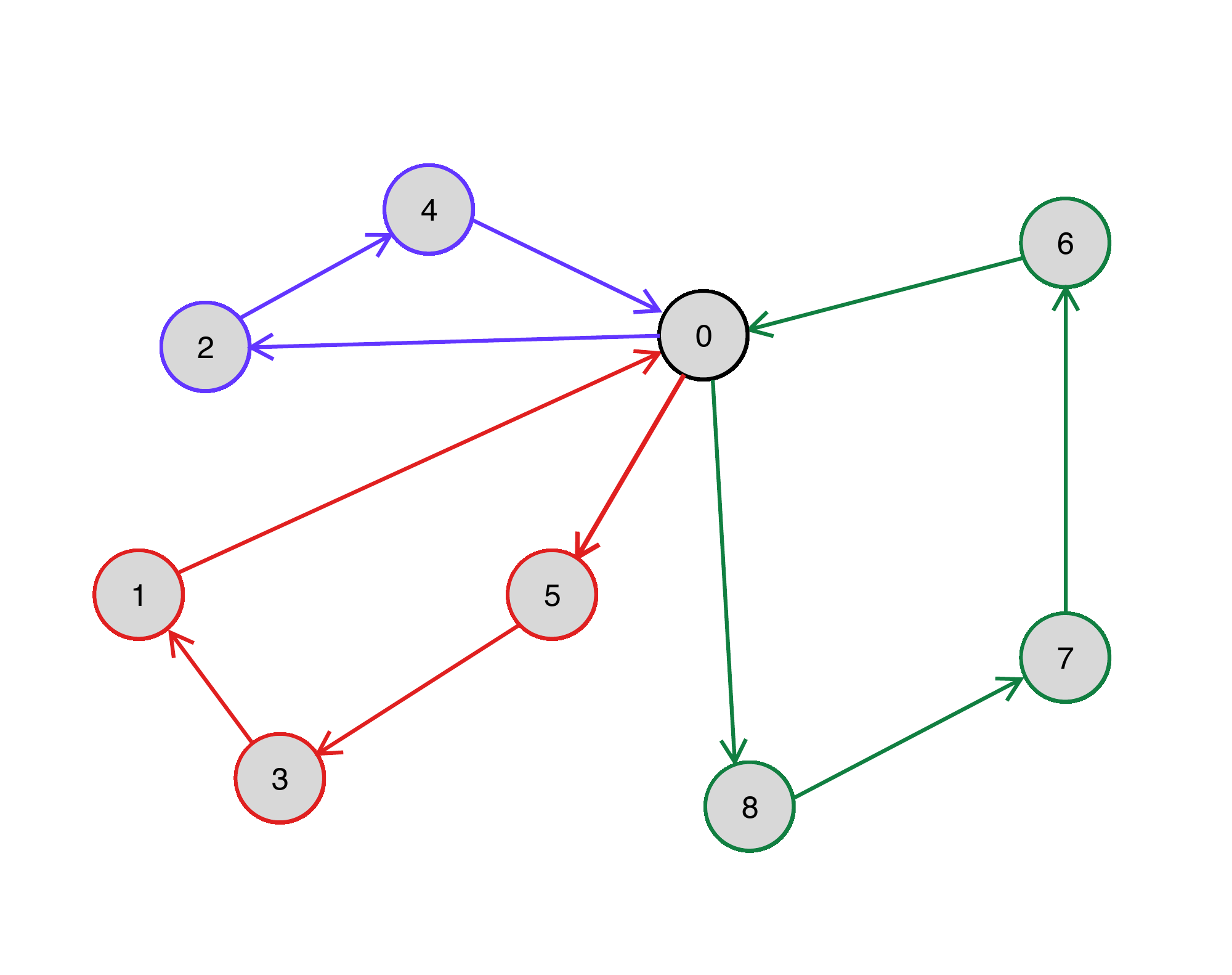}
        \caption{VRP routes of decoded vector.  The routes of vehicles with indices 9, 10, and 11 are colored, respectively, red, blue, and green.}
        \label{fig:VRP-routes}
    \end{subfigure}
    \caption{Example of decoder for VRP.}
    \label{fig:VRP}
\end{figure}

The main goal in producing a solution for a VRP is to assign customers to vehicles and sequence them on each vehicle. 
\citet{ResWer15a} describe a decoder for the VRP. In this decoder,
we assume there are $n$ customers and $m$ vehicles.  Solutions are encoded as vectors $\chi$ of $n+m$ random keys.  The first $n$ keys correspond to the
customers that should be visited, while the last $m$ keys correspond to the vehicles.
Decoding is accomplished by sorting the keys in $\chi$.  The last $m$ keys serve as the demarcation of customers assigned to the vehicles.
Sorted keys can be rotated so that the index of last key in $\pi(\sigma(\chi))$ is always a vehicle key.
Suppose the indices of the sorted vector of random keys appear as 
\begin{displaymath}
    \pi(\sigma(\chi')) = (\ldots, v, a, b, c, u, \ldots),
\end{displaymath}
where $v$ and $u$ are vehicles indices and $a, b$, and $c$ are customer indices. We assign customers $a, b$, and $c$ to vehicle $u$ and sequence the customers on that vehicle as $a$ to $b$ to $c$.
If vehicle key $v$ is followed immediately by vehicle key $u$, then vehicle $u$ is not assigned customers and is therefore not used.
Consider the example in Figure~\ref{fig:VRP-decoder} where we have eight customers and three vehicles.
Sorting the vector $\chi$ gives us the sorted vector $\sigma(\chi)$ and the corresponding indices of the sorted vector $\pi(\sigma(\chi))$ that encodes the assignment of customers to vehicles and their sequencing.
In this solution, the vehicle of key index 9 is assigned customers 5, 3, and 1, and vehicle 9 visits customer $5$, then customer $3$, and lastly
customer $1$.
The vehicle of key 10 is assigned customer 2, which it visits first, and then customer 4, visited last.  Lastly, customers 8, 7, and 6 are assigned to vehicle 11, which visits them in that order, i.e. 8 then 7, and finally customer 6.
The three routes are show in Figure~\ref{fig:VRP-routes}.

\section{Random-Key Optimizer}\label{sec:rko}

Random-Key Optimizer (RKO) is an efficient metaheuristic framework rooted in the concept of the random-key genetic algorithm (RKGA), initially proposed by \cite{Bean_RKGA_1994}. This approach encodes solutions as vectors of random keys — real numbers uniformly distributed in the interval [0,1) — enabling a unique representation of combinatorial optimization problems in a continuous search space. The RKGA evolves a population of random-key vectors over multiple generations, employing a problem-specific decoder to transform each vector into a feasible solution and evaluate its quality. This evolution process typically involves elitism, mutation, and crossover operations. The biased random-key genetic algorithm (BRKGA) \citep{Gonçalves_BRKGA_2011} refines this approach by introducing a bias towards elite solutions in the crossover phase. An up-to-date and complete review of the BRKGA is found in \cite{LONDE2024}.

RKGA and BRKGA represent a class of problem-independent metaheuristics for finding optimal or near-optimal solutions to optimization problems. These approaches operate indirectly in a continuous $n$-dimensional unit hypercube, employing a problem-specific decoder to map solutions from the continuous space to the discrete problem domain. This modular design allows the solver to be implemented once and reused to solve several problems by implementing a problem-specific decoder. Examples of Application Programming Interfaces (APIs) for BRKGA are \cite{TosRes15a}, \cite{andrade2021multi}, \cite{OliCarOliRes22a}, and \cite{Chaves_BRKGA_QL_2021}.

Some works extended the RKO framework to incorporate other metaheuristic paradigms.
\cite{lin2010efficient} and \cite{bewoor2017evolutionary, bewoor2018production} introduced a continuous Particle Swarm Optimization (PSO) algorithm integrated with a random-key encoding scheme to generate permutations.
\cite{garcia2015random} developed a Harmony Search (HS) heuristic utilizing random-key encoding to solve job-shop scheduling problems. In their approach, applying HS operators to random-key encoded harmonies consistently yielded feasible scheduling solutions.
\cite{ouaarab2015random} presented a random-key cuckoo search for the travelling salesman problem.
\cite{pessoa2018heuristics} presented four constructive heuristics for the flowshop scheduling problem, alongside BRKGA, Iterated Local Search (ILS), and Iterated Greedy Search (IGS) approaches that explore feasible sequence-based neighborhoods.
In related work, \cite{andrade2019scheduling} developed BRKGA and ILS approaches for the machine-dependent scheduling problem, utilizing a decoder that converts random-key vectors into schedules. This straightforward permutation-based decoder was later adapted for Tabu Search (TS) and Simulated Annealing (SA) implementations.
\cite{andrade2021multi} introduced Implicit Path-Relinking (IPR), which explores the separation between problem and solution spaces by constructing paths within the unit hypercube and using the decoder for solution evaluation.
\cite{Schuetz_RKO_2022} applied the RKO framework to robot motion planning, applying BRKGA and dual annealing to efficiently search the random-key space. In that paper the term RKO was proposed.
For the tree hub location problem, \cite{ManPolMacJulProGiaSalCha23a} deployed RKO implementations based on SA, ILS, and Variable Neighbourhood Search (VNS).
Expanding the metaheuristic portfolio, \cite{RK-GRASP} developed a GRASP-based RKO approach and demonstrated its effectiveness across diverse optimization challenges, including the traveling salesman problem, Steiner triple covering problem, node capacitated graph partitioning problem, and job sequencing with tool switching problem.

\begin{figure}[ht]
    \centering
    \includegraphics[width=0.95\linewidth]{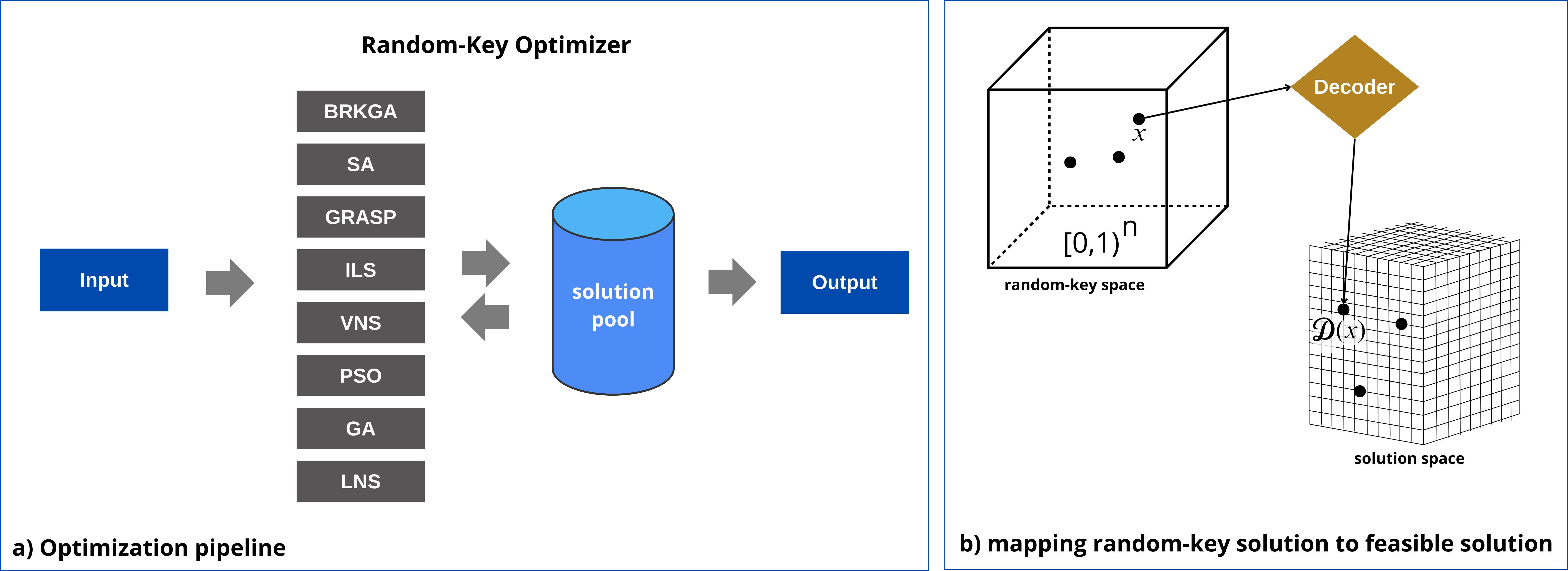}
    \caption{Schematic illustration of the RKO approach. Adapted from \cite{Schuetz_RKO_2022}.}
    \label{fig:RKO}
\end{figure}

Figure \ref{fig:RKO} presents a schematic representation of the RKO approach. In Figure \ref{fig:RKO}a, we can observe the optimization process that receives as input an instance of a combinatorial optimization problem and returns the best solution found during the search process. Diverse metaheuristics guide the optimization process, each employing distinct search paradigms while leveraging the random-key representation for solution encoding. These algorithms balance exploration and exploitation within the random-key space. In parallel computation, these metaheuristics engage in collaborative search by exchanging high-quality solutions through a shared elite solution pool. This pool, dynamically updated throughout the optimization process, is a repository of the most promising solutions, facilitating knowledge transfer across different search strategies and promoting convergence towards optimal solutions. Figure \ref{fig:RKO}b illustrates the mapping schema that links the random-key and solution space via the problem-dependent decoder. After this transformation, it is possible to evaluate the quality of the solutions.

The RKO framework demonstrates considerable flexibility, allowing for the adaptation of various metaheuristics beyond those currently implemented.  Researchers can potentially integrate additional population-based or single-solution metaheuristics, such as ant colony optimization or tabu search, by adapting their core mechanisms to operate within the random-key solution space. Furthermore, the elite solution pool inherent in the RKO framework presents an opportunity for implementing path-relinking strategies. Path-relinking \citep{glover2003scatter}, a method for generating new solutions by exploring trajectories connecting high-quality solutions, could be applied to pairs of solutions from the elite pool. This approach may enhance the framework's ability to intensify the search in promising regions of the solution space, potentially leading to improved solution quality or faster convergence.

\section{Framework}
\label{sec:framework}

This section presents the RKO framework developed to find optimal or near-optimal solutions to combinatorial optimization problems. First, we present the components of the RKO that are used by the metaheuristics: initial and pool of elite solutions (\Cref{sec:pool}), shaking (\Cref{sec:shaking}), blending (\Cref{sec:blending}), and local search (\Cref{sec:rvnd}). Then, we present the classical metaheuristics adapted to the random-key paradigm (\Cref{sec:MH}) and an online parameter control method (\Cref{sec:Qlearning}).

The RKO framework, including its source code in C++ and documentation, is freely available to researchers and practitioners through our public GitHub repository at \url{github.com/RKO-solver}.

\subsection{Initial and pool of elite solutions}
\label{sec:pool}

The metaheuristics initialize with solutions represented by $n$-dimensional vectors $\chi \in [0,1)^n$, where each key $x_i$ is randomly generated within the half-open interval $[0,1)$. The quality of each solution is quantified by an objective function value, obtained through the application of a problem-specific decoder function $\mathcal{D}$ to $\chi$, denoted as $f(\mathcal{D}(\chi))$.

The RKO framework maintains a shared pool of elite solutions ($\mathit{pool}$) accessible to all metaheuristics. This $\mathit{pool}$ is initialized with $\lambda$ randomly generated random-key vectors. To enhance the quality of the initial $\mathit{pool}$, each solution undergoes refinement via the Farey Local Search heuristic (detailed in \Cref{sec:farey}). To preserve diversity within the $\mathit{pool}$, any clone solutions (i.e., solutions with identical objective function values) are subjected to a perturbation process using the shaking method (described in \Cref{sec:shaking}).

Any solution generated by a metaheuristic that improves its current best solution is considered for inclusion in the elite $\mathit{pool}$. This inclusion is contingent upon the absence of an existing clone within the $\mathit{pool}$. Upon acceptance of a new solution, the $\mathit{pool}$ maintains its size constraint by eliminating the solution with the worst objective function value, thereby ensuring a continuous improvement in the overall quality of the elite $\mathit{pool}$.

\subsection{Shaking}
\label{sec:shaking}

To modify a random-key vector $\chi$, a perturbation rate \( \beta \in \mathbb{R}\) and $0 \leq \beta \leq 1$ is employed. This value is randomly generated within a specified interval \([\beta_{min}, \beta_{max}]\), which should be defined according to the specific metaheuristic approach being used. The shaking method, inspired by the approach proposed by \cite{ANDRADE201967}, applies random modifications to the random-key values by utilizing four distinct neighborhood moves:

\begin{itemize}
    \item \textit{Swap}: Swaps the value of two randomly selected random keys \( \chi_i \) and \( \chi_j \).
    \item \textit{Swap Neighbor}: Swaps the value of a randomly selected random key \( \chi_i \) with its neighbor \( \chi_{i+1} \).  In case $i=n$, then $\chi_{n}$ is swapped with $\chi_{1}$.
    \item \textit{Mirror}: Changes the value of a randomly selected random key \( \chi_i \) to its complement ($1-\chi_i$).
    \item \textit{Random}: Assigns a newly generated random value within the interval \([0, 1)\) to a randomly selected random key \( \chi_i \).
\end{itemize}

\begin{algorithm}[H]
\caption{Shaking method}
\label{alg:shaking}
\SetAlgoLined
\KwIn{Random-key vector $\chi$, $\beta_{min}$, $\beta_{max}$}
\KwOut{Changed random-key vector $\chi$}
Generate shaking rate $\beta$ randomly within the interval $[\beta_{min}, \beta_{max}]$\;
\For{$k \gets 1$ \textbf{to} $\beta \times n$}{
    Randomly select one shaking move $m$ from $\{1,2,3,4\}$\;
    \Switch{$m$}{
        \Case{$1$}{Apply Random move in $\chi$\;}
        \Case{$2$}{Apply Mirror move in $\chi$\;}
        \Case{$3$}{Apply Swap move in $\chi$\;}
        \Case{$4$}{Apply Swap Neighbor move in $\chi$\;}
    }
}
\Return $\chi$\;
\end{algorithm}

\Cref{alg:shaking} outlines the shaking method. A shaking rate $\beta$ is initially randomly generated (line 1). The main loop (lines 2-18) then iterates over the random-key vector $\chi$, applying perturbations $\beta \times n$ times. During each iteration, one of the four neighborhood moves is randomly selected and applied. Each move is performed in position of $\chi$ selected at random. After completing the perturbations, the modified random-key vector $\chi$ is returned (line 19). This vector must be decoded during the metaheuristics search process.

\subsection{Blending}
\label{sec:blending}

The blending method creates a new random-key vector by combining two parent solutions ($\chi^a$ and $\chi^b$). This process extends the uniform crossover (UX) concept \citep{davis1989handbook}, incorporating additional stochastic elements. For each position in the vector, a probability $\rho \in [0,1]$ determines whether the corresponding random key from $\chi^a$ or $\chi^b$ is inherited. We introduce a $\mathit{factor}$ parameter to modulate the contribution of $\chi^b$: when $\mathit{factor}=1$, the original key is used; when $\mathit{factor}=-1$, its complement ($1.0 - \chi^b_i$) is employed. Furthermore, with a small probability $\mu \in [0,1]$, the algorithm generates a novel random value within the $[0,1)$ interval, injecting additional diversity into the solution $\chi^c$. \Cref{alg:blending} presents the pseudocode of the blending method. Note that in our mathematical notation and pseudocode, all indices are 1-based.

\begin{algorithm}[htbp]
\caption{Blending method}
\label{alg:blending}
\SetAlgoLined
\KwIn{Random-key vector $\chi^a$, Random-key vector $\chi^b$, $\mathit{factor}$, $\rho$, $\mu$}
\KwOut{New random-key vector $\chi^c$}
\For{$i \gets 1$ \textbf{to} $n$}{
    \If{$\textrm{{\fontfamily{pcr}\selectfont UnifRand}}(0,1) < \mu$}{
        $\chi^c_i \leftarrow \textrm{{\fontfamily{pcr}\selectfont \textit{UnifRand}}}(0,1)$\;
    }
    \Else{
        \If{$\textrm{{\fontfamily{pcr}\selectfont UnifRand}}(0,1) < \rho$}{
            $\chi^c_i \leftarrow \chi^a_i$
        }
        \Else{
            \If{$\mathit{factor}=1$}{$\chi^c_i \leftarrow \chi^b_i$}
            \If{$\mathit{factor}=-1$}{$\chi^c_i \leftarrow 1.0 - \chi^b_i$}
        }        
    }
}
\Return $\chi^c$\;
\end{algorithm}

\subsection{Local Search}
\label{sec:rvnd}

We introduce the local search procedure used in the RKO algorithms, which is carried out by the Randomized Variable Neighborhood Descent (RVND) algorithm \citep{subramanian2010parallel}. The RVND is an extension of the Variable Neighborhood Descent (VND) method of \cite{Mladenovic_VNS_1997}. VND operates by exploring a finite set of neighborhood structures, denoted as \( N_k \) for \( k = 1, \ldots, k_{\max} \), where \( N_k(\chi) \) represents the set of solutions within the \( k \)-th neighborhood of a random-key vector \( \chi \). Unlike standard local search heuristics, which typically utilize a single neighborhood structure, VND leverages multiple structures to enhance the search process. The sequence in which these neighborhoods are explored is crucial to the effectiveness of VND. To address this challenge, RVND randomly determines the order of neighborhood heuristics applied in each iteration, thereby efficiently navigating diverse solution spaces. This approach is well-suited for random-key spaces, allowing users to either implement classic heuristics to the specific problem and encode the locally optimal solution into the random-key vector or employ random-key neighborhoods independent of the problem, using the decoder to iteratively refine the solution.

\Cref{alg:RVND} presents the pseudo-code for the RVND algorithm. The process begins by initializing a list of neighborhoods ($\mathit{NL}$) containing \( k_{\max} \) heuristics. The algorithm then iteratively selects a neighborhood \(\mathcal{N}^i\) at random and searches for the best neighboring solution (\(\chi'\)) within that neighborhood. If the new solution \(\chi'\) improves upon the current solution (\(\chi\)), \(\chi\) is updated to \(\chi'\), and the exploration of neighborhoods is restarted. If no improvement is found, the current neighborhood \(\mathcal{N}^i\) is removed from the list. This procedure continues until the list of neighborhoods is exhausted. Ultimately, the algorithm returns the best solution \(\chi\) found.

\begin{algorithm}[htbp]
    \KwIn{$\chi$}
    \KwOut{The best solution in the neighborhoods.}
    Initialize the Neighborhood List ($\mathit{NL}$)\;
    \While{$\mathit{NL} \neq 0$}{
        Choose a neighborhood $\mathcal{N}^i \in \mathit{NL}$ at random\;
        Find the best neighbor $\chi'$ of $\chi \in \mathcal{N}^i$\;
        \If{$f(\mathcal{D}(\chi')) < f(\mathcal{D}(\chi))$}{
            $\chi \leftarrow \chi'$\;
            Restart $\mathit{NL}$\;
        }
        \Else{
            Remove $\mathcal{N}^i$ from the $\mathit{NL}$ \;
        }
    }
    \Return $\chi$
\caption{RandomizedVND} \label{alg:RVND}
\end{algorithm}

We developed four problem-independent local search (LS) heuristics designed to operate within the random-key solution space. These heuristics constitute distinct neighborhood structures integrated into the RVND: Swap LS, Farey LS, Mirror LS, and Nelder-Mead LS. Each heuristic employs a unique strategy to exploit the solution landscape, enhancing the algorithm's capacity to find local optimal solutions.

\subsubsection{Swap Local Search}

The Swap LS heuristic involves swapping the positions of two random keys of the random-key vector. 
The process begins by generating a vector $RK$, which is a random permutation of the indices $\{1,2,\dots,n\}$ of the random keys. This ensures that the neighborhood is explored in a randomized order, as the sequence in which pairs of keys are swapped varies with each run.
The Swap LS procedure is detailed in \Cref{alg:swap}. The best solution found (\(\chi^{\mathit{best}}\)) in this neighborhood is updated in line 2. The main loop (lines 3-9) iteratively exchanges each pair of random keys at indices \( i \) and \( j \) in \( \mathit{RK} \). A first-improvement strategy is employed in each iteration, either continuing the search from the newly found best solution or reverting to the previous best solution. The best-found solution in this neighborhood is returned. 

\begin{algorithm}[htbp]
\caption{Swap Local Search}
\label{alg:swap}
\KwIn{Random-key vector $\chi$}
\KwOut{Best random-key vector $\chi^{best}$ found in the neighborhood}
Define a vector $\mathit{RK}$ with random order for the random-key indices\;
Update the best solution found $\chi^{best} \leftarrow \chi$\;
\For{$i \gets 1$ \textbf{to} $n-1$}{
    \For{$j \gets i+1$ \textbf{to} $n$}{
        Swap random keys $\mathit{RK}_i$ and $\mathit{RK}_j$ of $\chi$\;
        \If{$f(\mathcal{D}(\chi)) < f(\mathcal{D}(\chi^{\mathit{best}}))$}{
            $\chi^{\mathit{best}} \leftarrow \chi$\;
        }
        \Else{
            $\chi \leftarrow \chi^{\mathit{best}}$\;
        }
    }
}
\Return{$\chi^{\mathit{best}}$}\;
\end{algorithm}

\subsubsection{Farey Local Search}
\label{sec:farey}

The Farey LS heuristic adjusts the value of each random key by randomly selecting values between consecutive terms of the Farey sequence \citep{niven1991introduction}. The Farey sequence of order \(\eta \in \mathbb{N}^+\) includes all completely reduced fractions between 0 and 1 with denominators less than or equal to \(\eta\), arranged in ascending order. For our application, we use the Farey sequence of order 7:

$$F = \left\{ 
\frac{0}{1}, 
\frac{1}{7}, 
\frac{1}{6}, 
\frac{1}{5}, 
\frac{1}{4}, 
\frac{2}{7}, 
\frac{1}{3}, 
\frac{2}{5}, 
\frac{3}{7}, 
\frac{1}{2}, 
\frac{4}{7}, 
\frac{3}{5}, 
\frac{2}{3}, 
\frac{5}{7}, 
\frac{3}{4}, 
\frac{4}{5}, 
\frac{5}{6}, 
\frac{6}{7}, 
\frac{1}{1} 
\right\}$$

This sequence creates 18 intervals that are used to generate new random key values. In each iteration of this heuristic, the random keys are processed in a random order as specified by the \( \mathit{RK} \) vector and the first-improvement strategy is applied. The procedure is detailed in \Cref{alg:farey}.

\begin{algorithm}[htbp]
\caption{Farey Local Search}
\label{alg:farey}
\KwIn{Random-key vector $\chi$}
\KwOut{Best random-key vector $\chi^{\mathit{best}}$ found in the neighborhood}
Define a vector $\mathit{RK}$ with random order for the random-key indices\;
Update the best solution found $\chi^{\mathit{best}} \leftarrow \chi$\;
\For{$i \gets 1$ \textbf{to} $n$}{
    \For{$j \gets 1$ \textbf{to} $|F|-1$}{
        Set the value of the random key $\mathit{RK}_i$ of $\chi$ with $\textrm{{\fontfamily{pcr}\selectfont \textit{UnifRand}}}(F_j,F_{j+1})$\;
        \If{$f(\mathcal{D}(\chi)) < f(\mathcal{D}(\chi^{\mathit{best}}))$}{
            $\chi^{\mathit{best}} \leftarrow \chi$\;
        }
        \Else{
            $\chi \leftarrow \chi^{\mathit{best}}$\;
        }
    }
}
\Return{$\chi^{\mathit{best}}$}\;
\end{algorithm}

\subsubsection{Mirror Local Search}

The Mirror LS heuristic modifies the current value of each random key by inverting it. This heuristic utilizes the \( \mathit{RK} \) vector to generate a random order of indices. For each index in this sequence, the value $\chi_{\mathit{RK}_i}$ of the corresponding random key is replaced with its complementary value (\(1 - \chi_{\mathit{RK}_i}\)). The first-improvement strategy is applied during this process. The procedure is detailed in \Cref{alg:invert}.

\begin{algorithm}[htbp]
\caption{Mirror Local Search}
\label{alg:invert}
\KwIn{Random-key vector $\chi$}
\KwOut{Best random-key vector $\chi^{\mathit{best}}$ found in the neighborhood}
Define a vector $\mathit{RK}$ with random order for the random-key indices\;
Update the best solution found $\chi^{\mathit{best}} \leftarrow \chi$\;
\For{$i \gets 1$ \textbf{to} $n$}{
    Set the value of the random key $\mathit{RK}_i$ of $\chi$ with its complement\;
    \If{$f(\mathcal{D}(\chi)) < f(\mathcal{D}(\chi^{\mathit{best}}))$}{
        $\chi^{\mathit{best}} \leftarrow \chi$\;
    }
    \Else{
        $\chi \leftarrow \chi^{\mathit{best}}$\;
    }
}
\Return{$\chi^{\mathit{best}}$}\;
\end{algorithm}

\subsubsection{Nelder-Mead Local Search}

The Nelder-Mead algorithm, originally proposed by \cite{nelder1965simplex}, is a numerical optimization technique designed to compute the minimum of an objective function in a multidimensional space. This heuristic method, which relies on function value comparisons rather than derivatives, is widely employed in nonlinear optimization scenarios where gradient information is unavailable or computationally expensive to obtain. The algorithm initializes with a simplex of $k+1$ points in a $k$-dimensional space and iteratively refines the simplex through a series of geometric transformations. These transformations include reflection, expansion, contraction (internal and external), and shrinking, each aimed at improving the worst point of the simplex.

In our research, we developed an adapted Nelder-Mead LS heuristic with three solutions: $\chi^1$, $\chi^2$, and $\chi^3$. The first solution is a current solution of the RVND, while the remaining two are randomly selected from the pool of elite solutions discovered during the optimization process. These solutions are ranked according to their objective function values, with $\chi^1$ representing the best and $\chi^3$ the worst. Figure \ref{fig:NMexample} illustrates a simplex polyhedron and the five possible transformations in the Nelder-Mead LS.

\begin{figure}[htbp]
    \centering
    \includegraphics[width=0.95\textwidth]{NelderMead.pdf}
     \vspace*{10pt}
    \caption{Illustrative example of the simplex polyhedron and the five moves of the Nelder-Mead. Source: based on \cite{kolda2003optimization}}
    \label{fig:NMexample}
\end{figure}

\Cref{alg:neldermead} presents the procedure for our adapted Nelder-Mead LS. We employ the blending method (\Cref{sec:blending}) to generate new solutions, using $\rho = 0.5$ and $\mu = 0.02$ after exhaustive preliminary computational tests. The procedure initializes with a simplex comprising three solutions ($\chi^1,\chi^2,\chi^3$). The simplex's centroid ($\chi^0$) is computed using the blending method between $\chi^1$ and $\chi^2$. The algorithm then enters its main loop, which continues until a termination condition is met. Each iteration explores the random-key search space through a series of simplex transformations:

\begin{enumerate}
    \item Reflection: Compute $\chi^r = \text{Blending}(\chi^0, \chi^3, -1)$
    \item Expansion: If $\chi^r$ outperforms $\chi^1$, compute $\chi^e = \text{Blending}(\chi^r, \chi^0, -1)$
    \item Contraction: If neither reflection nor expansion improve $\chi^1$ or $\chi^2$: \\
            - Outside contraction: If $\chi^r$ is better than $\chi^3$, $\chi^c = \text{Blending}(\chi^r, \chi^0, 1)$\\
            - Inside contraction: Otherwise, $\chi^c = \text{Blending}(\chi^0, \chi^3, 1)$
    \item Shrinking: If contraction fails, the entire simplex contracts towards $\chi^1$: $\chi^i = \text{Blending}(\chi^1, \chi^i, 1)$ for $i=2,3$
\end{enumerate}

The algorithm terminates based on a predefined condition. In this study, we set the maximum number of iterations to $n \cdot e^{-2}$.

\begin{algorithm}[htbp]
    \KwData{$\chi^1$, $\chi^2$, $\chi^3$, $n$, $h$}
    \KwResult{The best solution found in \textit{simplex} $X$.}
    Initialize simplex: $X \leftarrow \{\chi^1, \chi^2, \chi^3\}$\;
    Sort simplex $X$ by objective function value\;
    Compute the simplex centroid $\chi^0 \leftarrow$ \texttt{Blending}$(\chi^1,\chi^2,1)$\;
    $\mathit{iter} \leftarrow 0$\; 
    $\mathit{numIter} \leftarrow n \cdot e^{-2}$\;
    \While{$\mathit{iter} < \textit{numIter}$}{
        $shrink \leftarrow 0$\; 
        $\mathit{iter} \leftarrow \mathit{iter} + 1$\; 
        Compute reflection solution 
        $\chi^r \leftarrow$  \texttt{Blending}$(\chi^0,\chi^3,-1)$\;
        \If{$f(\mathcal{D}(\chi^r)) < f(\mathcal{D}(\chi^1))$}{
            Compute expansion solution  $\chi^e \leftarrow$  \texttt{Blending}$(\chi^r,\chi^0,-1)$\;
            \If{$f(\mathcal{D}(\chi^e)) < f(\mathcal{D}(\chi^r))$}{
                $\chi^3 \leftarrow \chi^e$\;
            }
            \Else{
                $\chi^3 \leftarrow \chi^r$\;
            }
        }
        \Else{
            \If{$f(\mathcal{D}(\chi^r)) < f(\mathcal{D}(\chi^2))$}{
                $\chi^3 \leftarrow \chi^r$\;
            }
            \Else{
                \If{$f(\mathcal{D}(\chi^r)) < f(\mathcal{D}(\chi^3))$}{
                    Compute contraction solution  $\chi^c \leftarrow$  \texttt{Blending}$(\chi^r,\chi^0,1)$\;
                    \If{$f(\mathcal{D}(\chi^c)) < f(\mathcal{D}(\chi^r))$}{
                        $\chi^3 \leftarrow \chi^c$;\
                    }
                    \Else{
                        $shrink \leftarrow 1$\;
                    }
                }
                \Else{
                    Compute contraction solution  $\chi^c \leftarrow$  \texttt{Blending}$(\chi^0,\chi^3,1)$\;
                    \If{$f(\mathcal{D}(\chi^c)) < f(\mathcal{D}(\chi^3))$}{
                        $\chi^3 \leftarrow \chi^c$;\
                    }
                    \Else{
                        $shrink \leftarrow 1$\;
                    }
                }
            }
        }
        \If{$\mathit{shrink} = 1$}{
            Replace all solutions except the best $\chi^1$ with  $\chi^i \leftarrow$ \texttt{Blending}$(\chi^1,\chi^i,1), i=2,3$\;
        }
        Sort simplex $X$ by objective function value\;
        Compute the simplex centroid  $\chi^0 \leftarrow$ \texttt{Blending}$(\chi^1,\chi^2,1)$\;
    }
    \Return $\chi^1$ 
\caption{Nelder-Mead Local Search}\label{alg:neldermead}
\end{algorithm}

\subsection{Metaheuristics}
\label{sec:MH}

The versatility of the Random-Key Optimization (RKO) framework extends beyond its initial application in genetic algorithms, allowing for integration with a wide array of metaheuristics, each tailored to continuous optimization in the half-open unit hypercube. In this paper, we explore the adaptability of RKO by implementing a comprehensive framework that incorporates eight distinct metaheuristics. This diverse set includes the BRKGA, simulated annealing (SA), greedy randomized adaptive search procedure (GRASP), iterated local search (ILS), variable neighborhood search (VNS), particle swarm optimization (PSO), genetic algorithm (GA), and large neighborhood search (LNS). Each metaheuristic offers an approach to explore the search space while benefiting from RKO's random-key representation and decoding mechanism features. This integration demonstrates RKO's capacity to adapt to various optimization paradigms, providing a robust and flexible framework for addressing combinatorial optimization problems.

All metaheuristics start the search process from randomly generated random-key vectors.  After generating the initial solutions, each metaheuristic follows its search paradigm.

The RKO framework employs a predefined CPU time limit as its stopping criterion, ensuring all metaheuristics run with equivalent computational time. We adopted this approach in order to have a fair comparison among diverse optimization methods when executed on identical hardware architectures. Nevertheless, the framework allows alternative stopping criteria, such as a maximum number of objective function evaluations or a specified convergence threshold. 

In the context of RKO, the parameters of each metaheuristic need to be set using offline optimization (parameter tuning) or online optimization (parameter control) strategies. Parameter tuning focuses on finding optimal parameter values for a specific algorithm and problem instance and remains fixed throughout optimization. On the other hand, parameter control seeks to enhance performance by dynamically adjusting the parameters that control the behavior of the metaheuristics. 

In this framework, users can choose between parameter tuning and parameter control. In parameter tuning, users conduct an initial set of experiments to make informed decisions, or they can rely on automated tools such as iRace \citep{birattari2002racing}, paramILS \citep{Hutter2009}, or REVAC \citep{nannen2007efficient}. This process can be very complex in RKO due to the number of parameters and problem- and instance-dependence. To overcome these difficulties, we developed a reinforcement-learning-inspired parameter control based on the Q-Learning method \citep{Watkins1992} (see \Cref{sec:Qlearning}).

The following sections provide a brief overview of the metaheuristics developed in this paper. They outline the key principles and mechanisms underlying each approach and highlight the innovative aspects of the random-key solutions.

\subsubsection{BRKGA}
The Biased Random-Key Genetic Algorithm (BRKGA) \citep{Gonçalves_BRKGA_2011} is an evolutionary metaheuristic that extends the concept of RKGA. The algorithm maintains a population $p$ of random-key vectors, evolving them over multiple generations. BRKGA introduces a bias towards elite solutions in its selection process, distinguishing it from standard RKGA. The population is partitioned at each iteration into elite (with $p_e < p/2$ solutions) and non-elite sets. The next generation is formed by directly copying all elite solutions, introducing a small set ($p_m < p-p_e$) of new random solutions (mutants), and generating offspring through the parameterized uniform crossover \citep{spears1991virtues}. This crossover preferentially selects genes from the elite parent, controlled by an inheritance probability $\rho > 0.5$. A decoder transforms these random-key vectors into feasible solutions, evaluating their fitness. The new population becomes the current population. Whenever a new best solution is discovered within a generation, we apply the RVND heuristic (see \Cref{sec:rvnd}) in this solution. This intensification strategy aims to explore the promising region surrounding the best-found solution.

\subsubsection{GA}
We also implemented a standard Genetic Algorithm (GA) \citep{holland1992adaptation, de1989genetic} to work with the random-key representation. GA is a nature-inspired metaheuristic that emulates the principles of natural selection and genetic evolution to solve optimization problems. In its traditional form, GA operates on a population of candidate solutions, each encoded as a chromosome. The algorithm progresses through generations, applying genetic operators such as selection, crossover, and mutation to evolve the population towards better solutions. Our modified GA performs selection using the tournament method, applies crossover by combining random keys from selected parent solutions with the \textit{blending method}, and implements mutation by perturbing individual random keys (see \Cref{sec:blending}). The algorithm creates an entirely new population of offspring for each generation to replace the current population with elitism for population evolution. We apply the RVND heuristic (see \Cref{sec:rvnd}) to the best solution of the current population and insert this improved solution into the new population. The parameters of the GA are the population size ($p$), and the crossover ($p_c$) and mutation ($\mu$) probabilities. Parameter $p_c$  represents the likelihood that two solutions will exchange the random keys equally or that the parents will copy to the next population. The parameter $\mu$ represents the likelihood that the random key will be randomly altered during the crossover.

\subsubsection{SA}
Simulated Annealing (SA) \citep{Kirkpatrick1983} is a metaheuristic inspired by the annealing process in metallurgy, where controlled cooling of materials leads to more stable, low-energy states. In optimization contexts, SA navigates the solution space by iteratively perturbing the current solution and accepting or rejecting the new solution based on a probabilistic criterion. In the case of random-key representation, the \textit{shaking method} performs the perturbation process (see \Cref{sec:shaking}). SA begins with a high temperature ($T_0$), allowing for frequent acceptance of worse solutions to escape local optima. As the temperature gradually decreases according to a cooling schedule defined by the parameter $\alpha$, the algorithm becomes increasingly selective, converging towards high-quality solutions. The Metropolis criterion controls the probability of accepting a worse solution based on the difference between the objective function values of the current solution and the new perturbed solution, as well as the current temperature. This mechanism enables SA to balance exploration and exploitation effectively. The performance of SA is influenced by the initial temperature ($T_0$), the cooling rate ($\alpha$), and the number of iterations at each temperature level ($\mathit{SA}_{\mathit{max}}$). In our implementation, the RVND heuristic (see \Cref{sec:rvnd}) is applied before cooling the temperature. 

\subsubsection{GRASP}
\cite{RK-GRASP} adapted the Greedy Randomized Adaptive Search Procedure (GRASP) \citep{Feo_GRASP_1995} and  Continuous-GRASP \citep{HirMenParRes07a} to solve combinatorial optimization problems using random-key representation. This algorithm has two phases: a constructive phase and a local search phase. The constructive phase uses a line search strategy inspired by C-GRASP to generate new solutions, while the local search phase employs the RVND (\Cref{sec:rvnd}). Our GRASP iteratively improves solutions by generating random-key vectors, adjusting the grid parameter $h$, and using a simulated annealing acceptance criterion to decide whether to accept new solutions. 

The constructive phase is an iterative process that perturbs a given solution using semi-greedy moves. Each iteration uses the line search to find the best objective function value for each random key that is not fixed, randomly generating new random keys in the sub-intervals defined by $h$. A Restricted Candidate List (RCL) is then created, containing indices of random keys that produced solutions within a range defined by a randomly set parameter $\gamma \in [0,1]$. An index is randomly selected from the RCL, its corresponding random key is updated with the value found by the line search, and this random key is fixed. This process continues until all random keys have been fixed, balancing randomness and greediness in solution construction. Initially, the parameter $h$ is set to $h_s$, and each iteration without improvement of the current solution makes the grid more dense ($h=h/2$), up to an end grid density ($h_e$).

\subsubsection{ILS}

Iterated Local Searcch (ILS) \citep{Lourenco_ILS_2003} is a metaheuristic that alternates between intensification and diversification to explore the solution space effectively. In our implementation, we adapt the ILS framework to operate within the random-key representation paradigm. The algorithm begins with an initial solution encoded as a vector of random keys. It then enters its main loop, where it iteratively applies local search to reach a local optimum, followed by a perturbation mechanism to escape this local optimum. We consider two components in this process: the \textit{shaking} method (detailed in \Cref{sec:shaking}) and the RVND method (described in \Cref{sec:rvnd}). The shaking method is our perturbation mechanism, introducing controlled randomness to the current solution while preserving some of its structure. In each iteration, a parameter $\beta$ controls the intensity of the perturbation, generating a random value within the interval $[\beta_{min}, \beta_{max}]$. The RVND, on the other hand, is utilized within the local search phase to generate new candidate solutions by combining features of the current solution with other elite pool solutions. By operating on random-key vectors throughout the search process, our adapted ILS maintains the flexibility and problem-independence characteristic of RKO while benefiting from the exploration-exploitation balance inherent to the ILS.

\subsubsection{VNS}
Variable Neighborhood Search (VNS) \citep{Mladenovic_VNS_1997} is a metaheuristic similar to ILS that systematically leverages neighborhood changes to escape local optima. In our implementation, VNS begins with an initial solution encoded as a vector of random keys and iteratively applies the \textit{shaking} method (\Cref{sec:shaking}) and the RVND method (\Cref{sec:rvnd}). The shaking method perturbs the current solution with a randomly selected intensity, denoted by $\beta$, which is defined by the current neighborhood as $k \times \beta_{min}$. If a better solution is found, the search returns to the first neighborhood ($k = 1$); otherwise, it proceeds to the next neighborhood ($k=k+1$). A maximum neighborhood number ($k_{max}$) is predefined. After the shaking phase, the RVND procedure is applied, systematically exploring multiple heuristics in a randomized order.

\subsubsection{PSO}
Similar to BRKGA and GA, Particle Swarm Optimization (PSO) \citep{kennedy1995particle} is a population-based metaheuristic inspired by the social behavior of bird flocking. In PSO, a group of $p$ candidate solutions, known as particles, navigate the search space by adjusting their positions based on their own best-known position (\(P_{\text{best}}^i\)) and the swarm’s best-known position (\(G_{\text{best}}\)). We adapt PSO by representing each particle as a vector of random keys. In each generation, all particles are updated by calculating their current velocity \(V^i_j\) using the following equation:

\[
V^i_j = w \cdot V^i_j + c_1 \cdot r_1 \cdot (P^i_{\text{best}} - \chi^i_j) + c_2 \cdot r_2 \cdot (G_{\text{best}} - \chi^i_j)
\]
where \(c_1\), \(c_2\), and \(w\) are parameters, and \(r_1\) and \(r_2\) are random numbers uniformly distributed in the real interval \([0,1]\).

With these updated velocities, we adjust all random keys \(j\) of particle \(i\) ($\chi^i_j$) by adding the corresponding velocity \(V^i_j\) to the current value \(\chi^i_j\). The positions \(P_{\text{best}}^i\) and \(G_{\text{best}}\) are then updated with the new population. Additionally, the RVND heuristic (\Cref{sec:rvnd}) is applied to one randomly selected particle in each generation.

\subsubsection{LNS}

Large Neighborhood Search (LNS) \citep{ropke2006unified} is a metaheuristic optimization technique designed for solving combinatorial problems by iteratively destroying and repairing solutions, thereby enabling the exploration of an ample solution space. In each iteration, LNS partially deconstructs the current solution and then reconstructs it, with the potential for improvement. We adapted this approach to utilize the random-key representation, beginning with an initial random solution. The deconstruction phase involves randomly removing a portion of the random keys, with the intensity of this phase defined by a random value within the interval $[\beta_{min}, \beta_{max}]$. The repair phase is inspired by the Farey local search (\Cref{sec:farey}), where new random values are generated for each removed random key within the intervals of the Farey sequence. The random key is then assigned the value that yields the best objective function result. This process continues until all random keys that have been removed have been repaired. The acceptance criterion in LNS is based on the Metropolis criterion, where worse solutions may be occasionally accepted depending on the current temperature, allowing the algorithm to escape local optima and explore different regions of the solution space. The process begins with an initial temperature ($T_0$), gradually reducing by a cooling factor $\alpha$ at each iteration. Additionally, whenever LNS identifies a better solution, a local search is performed using the RVND method (\Cref{sec:rvnd}).

\subsection{Q-Learning}
\label{sec:Qlearning}

Q-Learning \citep{Watkins1992} is a classical heuristic algorithm designed to seek the solution to stochastic sequential decision problems, which can be modelled by means of Markov Decision Processes (MDP) \citep{Puterman2014}. The goal is to find an optimal stationary policy that minimizes the long-term cost by simulating the system's transitions under a greedy policy that is refined via stochastic approximation \citep{Robbins-Monro1951}. Under appropriate parameter control conditions, the algorithm is guaranteed to converge to an optimal solution as long as each action is tried infinitely often for each possible system's state.

To guide the parameter exploration, we define a Markov decision process for each metaheuristic, with each state $s$ in the state space $S$ representing a possible parameter configuration and each action $a \in A(s)$ representing a transition between parameter configurations available in $s$. $$A= \displaystyle \bigcup_{s \in S} A(s)$$ is the set of feasible actions in $S$ \citep[e.g.,][]{Puterman2014}. We assume that the user prescribes a discrete set of configurations to comprise the state space $S$, and that each action corresponds to a change in a single parameter while the remaining parameters are kept unaltered. We utilize $Q-$Learning to navigate the state space and optimize the parameter choice.

Our approach was built upon the BRKGA-QL framework \citep{Chaves_BRKGA_QL_2021} and introduced a novel MDP representation. In the BRKGA-QL, each state corresponds to a parameter, and the actions represent the possible values for those parameters. In contrast, the new MDP representation is more efficient and robust, offering enhanced performance and stability compared to the previous version. 

At each step $t$, the agent selects an action $a_t \in A(s_t)$ from the set of feasible actions in the current state $s_t$ using the $Q$-Table and following an $\epsilon$-greedy policy \citep[e.g,][]{Sutton1998}. The objective is to maximize the total reward by choosing the action with the highest $Q$-value in state $s_t$ with a probability of $1 - \epsilon$, while with a probability of $\epsilon$, a random action is selected to encourage exploration. This action leads to a new parameter configuration.

Next, the parameters are adjusted, and a new iteration of the metaheuristic is executed, after which the agent receives a reward $R_{t+1}$, either positive or negative. The $Q$-Table is updated, and the agent transitions to the new state $s_{t+1}$. The following function determines the reward $R_{t+1}$:

$$
R_{t+1} = \left\{\begin{matrix}
1, & \mbox{if  } f^{t+1}_b < f^{t}_b
\\ 
\frac{f^{t}_{b} - f^{t+1}_b}{f^{t+1}_b}, & \textrm{otherwise}
\end{matrix}\right.
$$

\noindent where $f^t_b$ is the best objective function value of the solutions in iteration (epoch) $t$.

We utilize a warm restart decay strategy for $\epsilon$ to balance exploration and exploitation \citep{chaves2024parallel}. This method resets the $\epsilon$ value at regular intervals. Specifically, for every $T$ computing period, where $T$ represents 10\% of the maximum computational time set by the user, we reset $\epsilon$. We apply a cosine annealing function at each step to control the decay of the $\epsilon$ parameter: 

$$
\epsilon^i = \epsilon_{min} + \frac{1}{2} (\epsilon^{i}_{max} - \epsilon_{min}) \left(1 + \cos\left (\frac{T^i_{cur}}{T} \pi \right )\right)
$$

\noindent where $\epsilon_{min} = 0.1$ and $\epsilon^i_{max} = \{1, 0.9, 0.8, 0.7, 0.6, 0.5, 0.4, 0.3, 0.2, 0.1\}$ define the range for $\epsilon$, while $T^i_{cur}$ tracks the elapsed time since the last reset.

The function $Q(s_t, a_t)$ represents the value of taking action $a_t$ in state $s_t$, indicating how effective this choice is in optimizing the expected cumulative reward. The update rule is given by the Bellmann equation \citep{bellman1952theory}:

$$
Q(s_t,a_t) = Q(s_t,a_t) + \mathit{lf} \left [ R_{t+1} + \mathit{df} \times \underset{a \in A(s_{t+1})}{\max} \ Q(s_{t+1},a) - Q(s_t,a_t) \right ].
$$

The value $Q(s_t, a_t)$ is updated in every iteration based on the immediate reward received. The value increases when the action leads to a positive reward $R_{t+1}$, and the maximum $Q$-value of the next state exceeds the current $Q(s_t, a_t)$, showing that the action was more valuable than previously estimated. Conversely, it decreases when receiving negative rewards, reflecting the undesirable outcomes of those actions.

The learning factor ($\mathit{lf}$) and discount factor ($\mathit{df}$) are parameters in $Q$-Learning, adjusted through deterministic rules during the search process. The $\mathit{lf}$ is updated based on runtime, prioritizing newly acquired knowledge in the early generations and gradually increasing the importance of the knowledge in the $Q$-Table ($\mathit{lf} = 1 - 0.9 \times \text{percentage of running time}$). The discount factor controls the weight of future rewards, with a value between 0 and 1. A higher $\mathit{df}$ (close to 1) emphasizes long-term rewards, while a lower value focuses on immediate gains. We adopt the same strategy of \cite{chaves2024parallel}, where $\mathit{df}$ is set to 0.8 to balance current and future rewards.

All metaheuristics can implement the adaptive control of parameters using the Q-Learning. The proposed approach follows a systematic flow in three phases: initially, the possible states and actions are defined, and the Q-Table is initialized with arbitrary values; during the search process, at the beginning of each iteration, an action is selected through the $\epsilon$-greedy policy, which determines the parameter values to be used; at the end of each iteration, the algorithm calculates the reward based on the obtained performance, updates the $Q$-Table values through the Bellman equation, and transitions to the next state. This mechanism allows the algorithm to progressively learn which parameter configurations are most effective in different states of the search process, dynamically adapting to the problem characteristics and search evolution.

\section{Applications}\label{sec:aplications}

This section presents the applications of the RKO framework to three combinatorial optimization problems classified as NP-hard:

\begin{itemize}
    \item $\alpha$-Neighbor $p$-Median Problem
    \item Node Capacitated Graph Partitioning Problem
    \item Tree Hub Location Problem
\end{itemize}

The RKO framework was coded in C++ and compiled with GCC using the O3 and OpenMP directives. All experiments were conducted on a computer with a Dual Xeon Silver 4114 20c/40t 2.2 GHz processor, 96GB of DDR4 RAM, running CentOS 8.0 x64. In the parallel RKO framework runs, each of the metaheuristics was executed in a separate thread. For the experiments involving individual metaheuristics, each was run in a single thread. Each instance was run five times per method, with a specific time limit as the stopping criterion. All instance sets and detailed results are available online at \url{https://github.com/RKO-solver}.

In the following subsections, we briefly describe the combinatorial optimization problems, the decoder process, and a summary of the computational experiments. For each problem, we present a table with the results found by the state-of-the-art (STOA) methods, the RKO framework, and each metaheuristic of the RKO run by itself. We use the prefix ``RKO-'' to identify the RKO-based metaheuristic algorithms. For each method and problem, we present an average of objective function values of the best solutions found in all instances (column $\mathit{Best}$), an average time, in seconds, in which the best solution was found (column best found at (s)), and the number of best-known solutions found (column $\#\mathit{BKS}$). We also calculate the relative percentage deviation ($\mathit{RPD}$) for each run and instance, defined by 

\[
\mathit{RPD} (\%) = \frac{(Z - \mathit{BKS})}{\mathit{BKS}} \times 100,
\]
where $Z$ is the solution value found by a specific method, and $\mathit{BKS}$ is the value of the best-known solution. We present an average of the $\mathit{RPD}$ for the best solution found in five runs (column $\mathit{RPD}_{\mathit{best}}$) and an average of all $\mathit{RPD}$s computed over all runs (column $\mathit{RPD}_{\mathit{avg}}$).

The parameters of the metaheuristics were tuned for each combinatorial optimization problem using an offline parameter tuning strategy based on the design of experiments (DoE). The values of the different parameters were fixed before executing the metaheuristics, with some cases employing adaptive deterministic rules. We utilized an experimental design approach, considering the parameters and their potential values, to identify the ``best'' value for each parameter through many experiments on a subset of the problem instances. The values used by RKO in each optimization problem are presented in the following.

\subsection{$\alpha$-Neighbor $p$-Median Problem}\label{sec:aNpMP}

The $\alpha$-Neighbor $p$-Median Problem ($\alpha$NpMP) \citep{Panteli2021} is a variant of the facility location problem where a predefined number, $p$, of facilities must be established, and each demand point is assigned to its nearest $\alpha$ facilities. This problem arises from the consideration that a facility may eventually close, yet the system or service must continue operating by reassigning demand points to other facilities. The objective of the $\alpha$NpMP is to minimize the total distance from each demand point to its $\alpha$ assigned facilities.

The $\alpha$NpMP is mathematically defined as follows: Given an undirected, weighted, and connected graph $G = (V, E)$ with $|V| = n$ vertices and $|E| = m$ edges, where each edge $(i,j) \in E$ has an associated non-negative weight $d_{ij} \in \mathbb{R}^+$, the $\alpha$NpMP seeks to find a subset $S \subseteq V$ of $p$ facilities $(1 \leq p \leq n)$ that minimizes the sum of the $\alpha$-median distances for all vertices. The $\alpha$-median distance of a vertex $i$, given a set of facilities $S$, is defined as $d_m(i,S,p) = \sum_i \min\{d_{ij} : j \in S', S' \subset S, |S'| = \alpha, \alpha \leq p\}$. This value represents the sum of the distances from vertex $i$ to its $\alpha$ closest facilities within $S$. The objective function of the $\alpha$NpMP can then be expressed as $\min \sum_{i \in V} d_m(i,S, p)$, subject to $|S| = p$ and $S \subseteq V$. In this problem, all vertices $i \in V$, including those selected as facilities, are assigned to their $\alpha$ nearest medians among the $p$ open facilities.

The $\alpha$NpMP has been addressed in two papers. \citet{Panteli2021} proposed a relaxation of the assignment constraint in the $p$-median problem, requiring each vertex to be allocated to its $\alpha$ nearest medians. To solve this problem, the authors developed the Biclustering Multiple Median heuristic (BIMM). Subsequently, \citet{chagas2024parallel} presented a mathematical model solvable by the Gurobi optimizer and proposed a basic parallel Variable Neighborhood Search (BP-VNS) algorithm. The BP-VNS successfully identified all optimal solutions, as Gurobi proved. In our study, we compare the performance of the RKO algorithm against these two heuristics (BIMM and BP-VNS) and the commercial solver.

We encoded solutions of the $\alpha$NpMP as a vector of random keys ($\chi$) of size $p$, where each key corresponds to a facility that will be opened. The decoding begins by generating a candidate list ($C$) containing all possible facilities. For each random key $i$, an index $k$ is selected from the candidate list based on the key's value using the formula $\left\lfloor \chi_i \times |C| \right\rfloor$, where for $r \in \mathbb{R}$, $\left\lfloor r \right\rfloor$ is the largest integer less than $r$. The facility corresponding to this index in $C$ is then added to the set of open facilities, and position $k$ is removed from the candidate list. This process repeats until $p$ facilities have been selected.

Once the open facilities are determined from the random key vector, the objective function value of the solution is calculated. For each vertex $j$ in the graph, the $\alpha$ closest open facilities are assigned to it. The objective function is the sum of the distances between each vertex and its $\alpha$ closest facilities.

\begin{figure}[hbtp]
    \centering
    \includegraphics[width=0.45\linewidth]{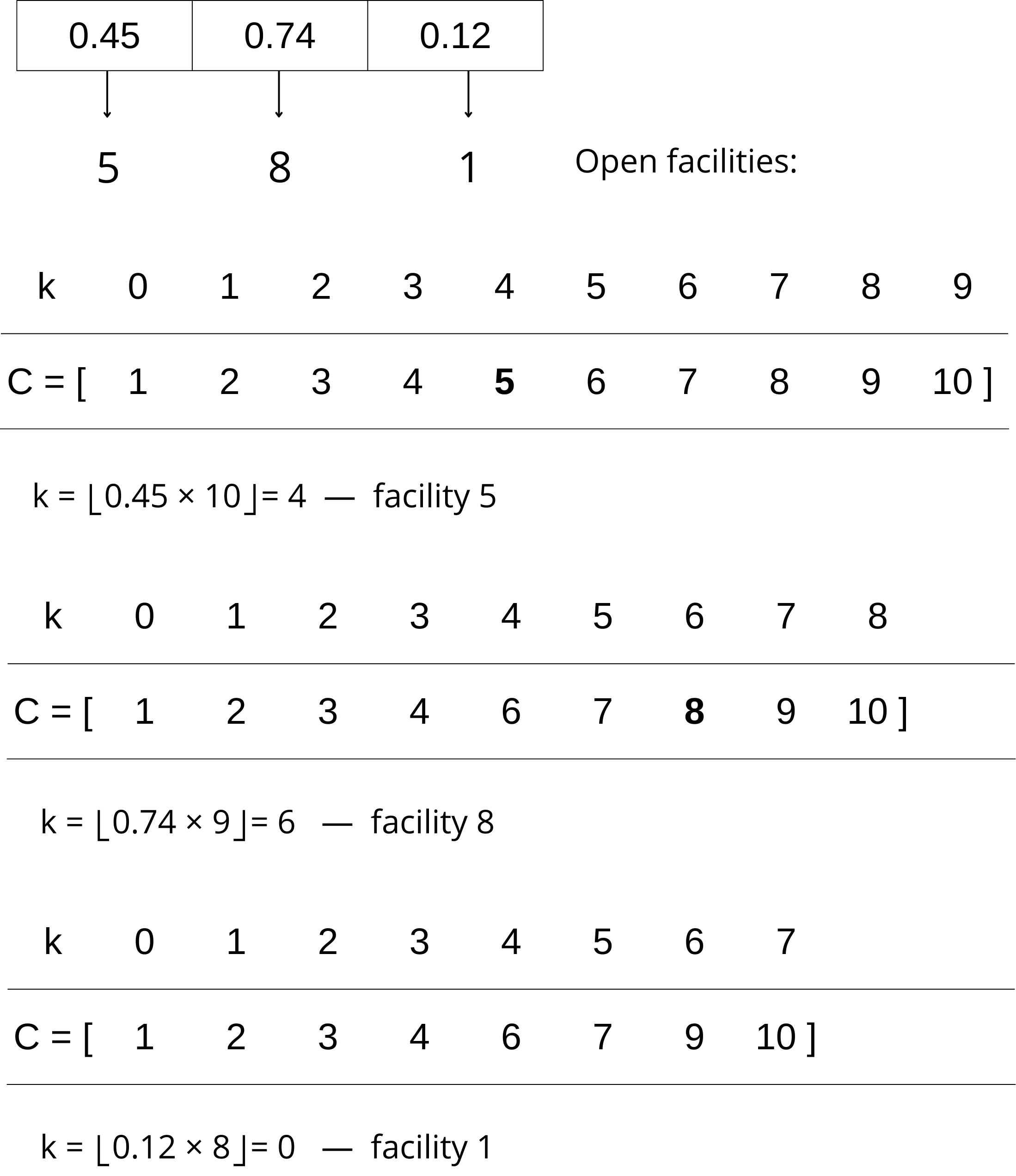}
    \caption{Example of the $\alpha$NpMP decoder with ten vertices and three facilities.}
    \label{fig:decoderANpMP}
\end{figure}

Figure \ref{fig:decoderANpMP} illustrates an example of the $\alpha$NpMP decoder with ten vertices and three facilities ($p=3$). We begin by defining the list $C$ of candidate facilities and then map a random-key vector directly into an $\alpha$NpMP solution. The first random key (0.45) corresponds to index 4 of the list $C$ ($k = \left\lfloor 0.45 \times 10 \right\rfloor = 4$), so facility 5 is opened and removed from the list $C$. The second random key (0.74) points to index 6 ($k = \left\lfloor 0.74 \times 9 \right\rfloor = 6$), representing facility 8, which is then opened and removed from the list $C$. Finally, the last random key (0.12) corresponds to index 0 in the list $C$ ($k = \left\lfloor 0.12 \times 8 \right\rfloor = 0)$, which is facility 1. Facility 1 is opened, completing the process. The open facilities are, therefore, vertices 5, 8, and 1.

We used instances from the OR-library \citep{beasley1990or} for the $\alpha$NpMP experiments. This set contains 40 instances with sizes ranging from $100$ to $900$ vertices. The number of facilities and $\alpha$ were set as $\{10,20\}$ and $\{5,10\}$, respectively. We have computational experiments for 80 instances. We compared the RKO results with Gurobi version 10.0.3, BIMM \citep{Panteli2021}, and  VNS \citep{chagas2024parallel}. We used a stopping criterion for the RKO methods based on the computational time equal to 10\% of the number of vertices for each instance (in seconds).

Table \ref{tab:parameterAlphaNpMP} presents the parameters of each metaheuristic used by the RKO to solve the $\alpha$NpMP.

\begin{table}[ht]
\caption{Parameters of the RKO metaheuristics to solve the $\alpha$NpMP.}
\label{tab:parameterAlphaNpMP}
\centering
\tiny
\begin{tabularx}{\textwidth}{lXrrrrrrrr}  \toprule
Parameters & Definition & BRKGA & GA & SA & ILS & VNS & GRASP & PSO & LNS \\
\midrule
$p$ & population size & 1597 & 1000 &  &  &  &  & 100 &  \\ 
$p_e$ & elite set & 0.10 &  &  &  &  &  &  &  \\ 
$p_m$ & mutant set & 0.20 &  &  &  &  &  &  &  \\ 
$\rho$ & inherit probability & 0.70 &  &  &  &  &  &  &  \\ 
$p_c$ & crossover probability &  & 0.85 &  &  &  &  &  &  \\ 
$\mu$ & mutation probability &  & 0.03 &  &  &  &  &  &  \\ 
$T_0$ & initial temperature &  &  & 10000 &  &  &  &  & 1000 \\ 
$\mathit{SA}_{\mathit{max}}$ & number of iterations &  &  & 100 &  &  &  &  &  \\ 
$\alpha$ & cooling rate &  &  & 0.99 &  &  &  &  & 0.90 \\ 
$\beta_{min}$ & minimum rate of shaking &  &  & 0.10 & 0.15 & 0.05 &  &  & 0.10 \\ 
$\beta_{max}$ & maximum rate of shaking &  &  & 0.20 & 0.40 &  &  &  & 0.30 \\ 
$k_{max}$ & number of neighborhoods &  &  &  &  & 6 &  &  &  \\ 
$h_s$ & start grid dense &  &  &  &  &  & 0.125 &  &  \\ 
$h_e$ & end grid dense &  &  &  &  &  & 0.00012 &  &  \\ 
$c_1$ & cognitive coefficient &  &  &  &  &  &  & 2.05 &  \\ 
$c_2$ & social coefficient &  &  &  &  &  &  & 2.05 &  \\ 
$w$ & inertia weight &  &  &  &  &  &  & 0.73 &  \\ \bottomrule
\end{tabularx}
\end{table}

Table \ref{tab:resultsAlphaNpMP} summarizes the results of the tests on the OR-Library instances. This table shows the name of the method, the average of the best solution found on the 80 instances, the $\mathit{RPD}$ of the best solution found in five runs and an average of the $\mathit{RPD}$s, the average of the best time to find the best solution in each run, and the number of $\mathit{BKS}$ found. A literature review shows that Gurobi found the optimal solution for all 80 instances tested. VNS \citep{chagas2024parallel} also achieved these optimal solutions, whereas BIMM \citep{Panteli2021} failed to find optimal solutions for all instances, with a $\mathit{RPD}_{\mathit{best}}$ of 2.55\%. In the case of RKO, the individual metaheuristics produced results close to the optimum, with RKO-VNS finding the optimal solution in 55 instances. However, when RKO was executed with all eight metaheuristics running in parallel and exchanging information through the elite pool, it found the optimal solution in 65 instances and produced results even closer to the optimum in all runs, with a $\mathit{RPD}_{\mathit{avg}}$ of 0.02.

\begin{table}[ht]
\centering
\caption{Summary of the $\alpha$NpMP results for the OR-library instances. }
\label{tab:resultsAlphaNpMP}
\begin{tabular}{lrrrrr} \toprule
Method               & \textit{Best}     & $\mathit{RPD_{best}}$ & $\mathit{RPD_{avg}}$  & best found at (s) & $\#\mathit{BKS}$ \\ \midrule
Gurobi$^2$                      & 84296.53 & 0.00     & -    & 155.67            & \textbf{80}    \\
BiMM$^1$  & 86465.14 & 2.55     & -    & 2.43              & 0     \\
BP-VNS$^2$                         & 84296.53 & 0.00     & -    & 0.38              & \textbf{80}    \\
RKO                        & 84300.39 & 0.003    & 0.02 & 28.70             & \textbf{65}    \\
RKO-BRKGA                  & 84310.84 & 0.01     & 0.07 & 23.11             & 54    \\
RKO-SA                     & 84349.73 & 0.04     & 0.10 & 38.67             & 51    \\
RKO-GRASP                  & 84328.86 & 0.03     & 0.10 & 41.02             & 51    \\
RKO-ILS                    & 84321.18 & 0.02     & 0.06 & 44.73             & 53    \\
RKO-VNS                    & 84321.68 & 0.02     & 0.06 & 42.43             & 55    \\
RKO-PSO                    & 84356.23 & 0.05     & 0.16 & 35.34             & 39    \\
RKO-GA                     & 84312.56 & 0.01     & 0.04 & 33.74             & 55    \\
RKO-LNS                    & 84335.30 & 0.03     & 0.12 & 38.84             & 50   
\\ \bottomrule
\multicolumn{6}{l}{$^1$ \citep{Panteli2021} $^2$ \citep{chagas2024parallel}}
\end{tabular}
\end{table}

To statistically evaluate the differences between the RKO method and its metaheuristics, we employed the Wilcoxon signed-rank test \citep{Wilcoxon}, a non-parametric test for comparing two related samples. We performed a pairwise comparison for all pairs of methods. Specifically, a one-sided test (Row $<$ Column) was used to determine if the performance of one method was statistically better than the other. The null hypothesis states that there is no difference in performance between the two methods. In contrast, the alternative hypothesis suggests that the method in the ``Row'' consistently achieved better performance ranks than the method in the ``Column''.

\begin{figure}[ht]
    \centering
    \includegraphics[width=1.0\linewidth]{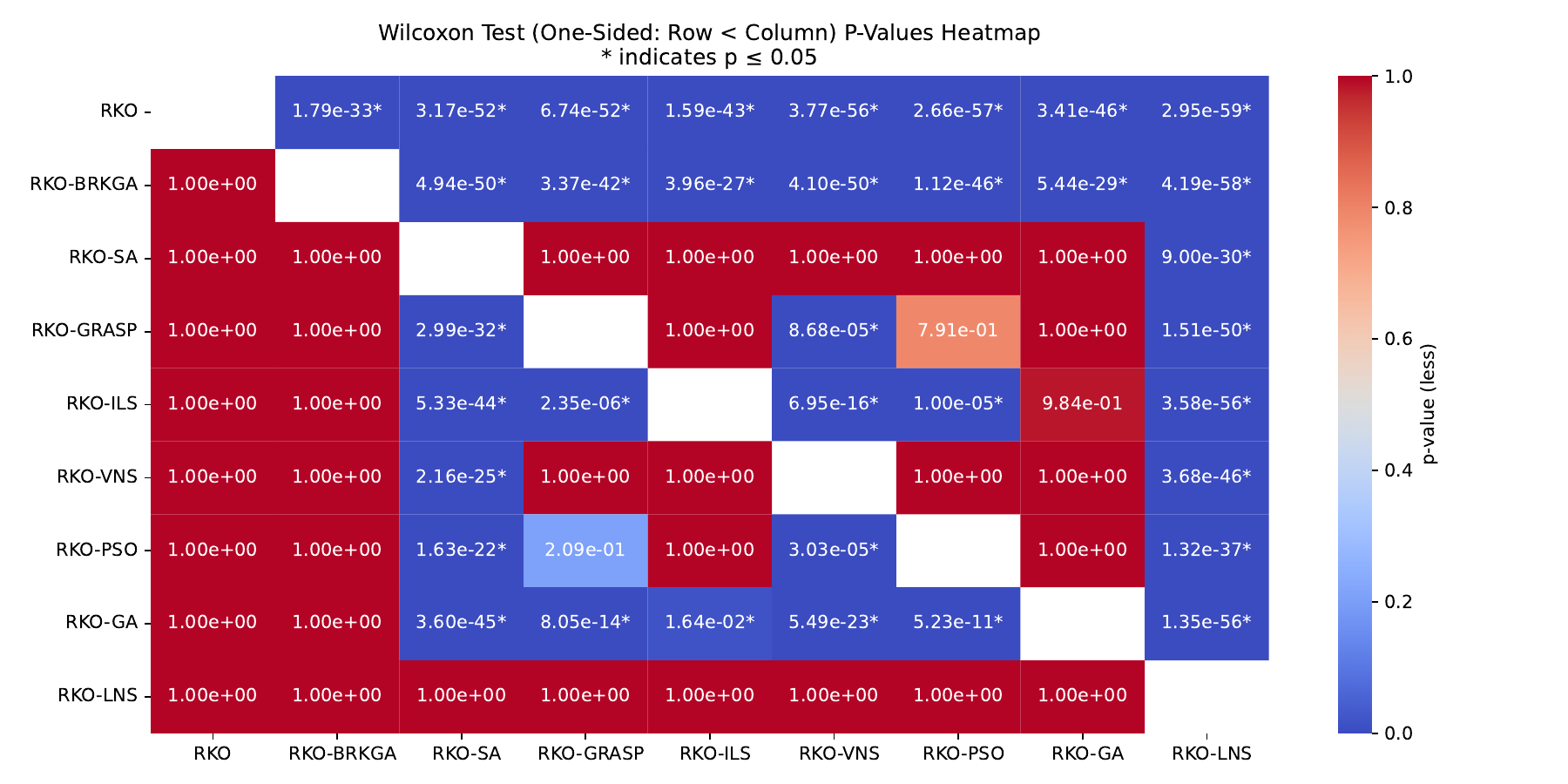}
    \caption{Heat map of the results ($p$-values) of the Wilcoxon signed-rank test for the RKO methods to solve the $\alpha$NpMP.}
    \label{fig:wilcoxonAlphaNpMP}
\end{figure}

The results of the Wilcoxon signed-rank test are presented in the heat map shown in Figure \ref{fig:wilcoxonAlphaNpMP}. The blue cells with an asterisk (*) represent cases with a statistically significant difference at the 95\% confidence level. The Wilcoxon test indicated that the RKO method produced significantly better results than all the individual metaheuristics with the random-key representation. However, no statistically significant differences were found when comparing GRASP and PSO, indicating a certain similarity in the performance of these methods.

The computational time performance of the RKO methods was evaluated using the performance profile technique introduced by \cite{dolan2002benchmarking}. The performance ratio for each method-instance pair is defined as:

\[
r_{i,h} = \frac{t_{i,h}}{\min\{t_{i,h} | h \in H\}},
\]

\noindent where $t_{i,h}$ is the average computational time for method $h$ to find the best solution for instance $i$, and $H$ is the set of all methods. To evaluate the quality of the results, a tolerance threshold was set at $\mathit{RPD}_{\mathit{best}} = 1\%$ for each instance. If the $\mathit{RPD}_{\mathit{best}}$ of a method's solution exceeded 1\%, its computational time for that instance was set to $\infty$, indicating a failure to satisfy the convergence criteria.

The cumulative distribution function $\rho_h(\tau)$ computes the probability that a method $h$ achieves a performance ratio $r_{i,h}$ within a factor $\tau$ of the best possible ratio. This is calculated as:

\[
\rho_h(\tau) = \frac{|\{i \in I : r_{i,h} \leq \tau\}|}{|I|},
\]

\noindent where $I$ is the set of all instances.

\begin{figure}[ht]
    \centering
    \includegraphics[width=0.8\linewidth]{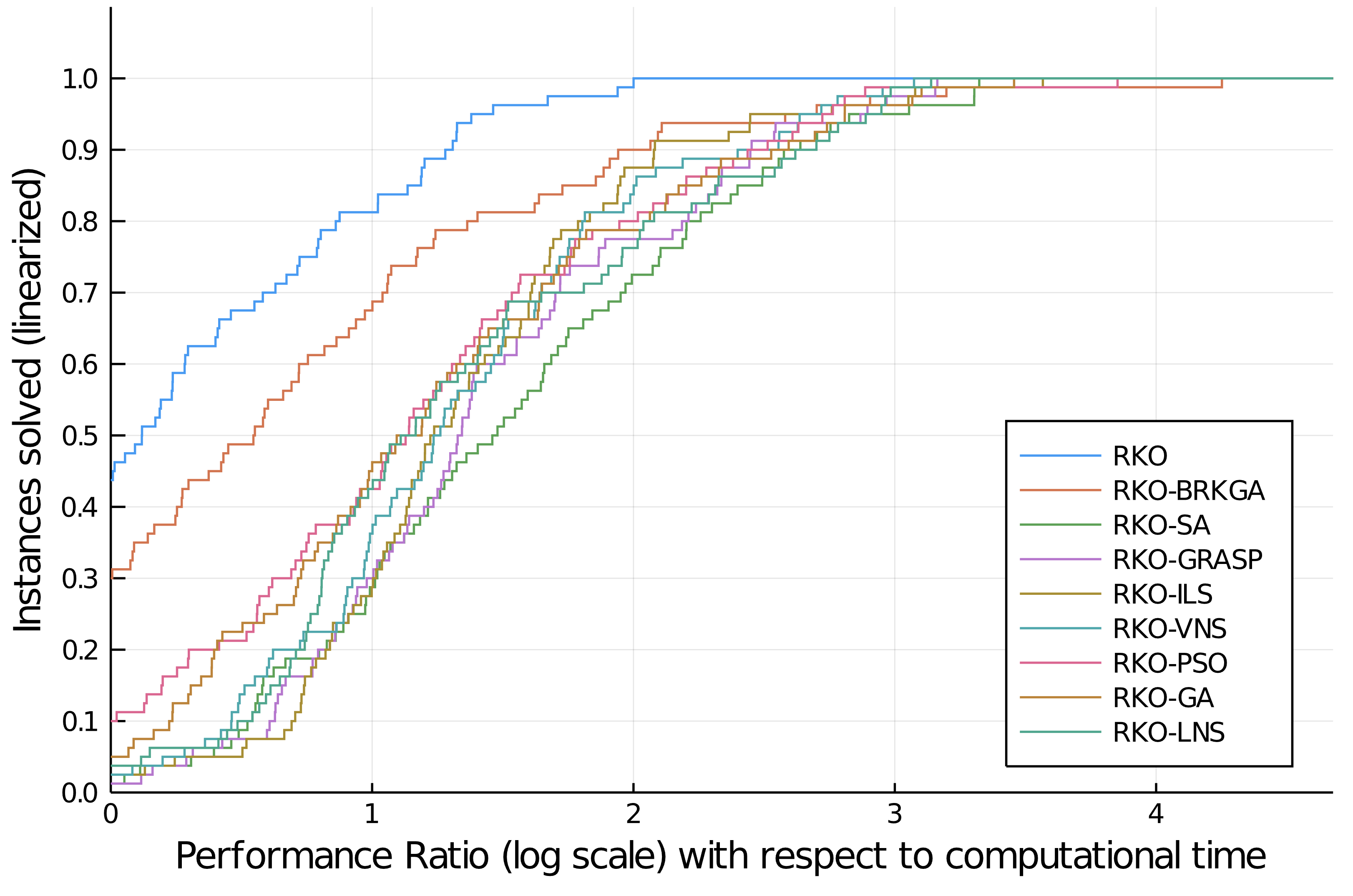}
    \caption{Performance profile of runtime for RKO methods to solve the $\alpha$NpMP.}
    \label{fig:perfprofAlphaNpMP}
\end{figure}

Figure \ref{fig:perfprofAlphaNpMP} presents the performance profiles of the RKO methods on a $\log_2$ scale. Using a performance profile with a convergence test, we can evaluate the accuracy of the methods. The RKO method exhibited the most robust results in terms of both computational time and solution quality, outperforming the individual metaheuristics on the OR-library instances. For the 1\% accuracy target ($\mathit{RPD}_{\mathit{best}} \leq 1$), RKO was able to find the target solution for all instances with a performance ratio of $\tau = 4$ (corresponding to $\log_2(\tau) = 2$). In contrast, the individual metaheuristics achieved 100\% at a performance ratio higher than $\tau = 8$ ($\log_2(\tau) = 3$). Additionally, RKO was the fastest solver for 45\% of the instances, while RKO-BRKGA was the fastest for 31\% and the other methods for less than 10\%. Looking at a performance ratio of $\tau = 2$ ($\log_2(\tau) = 1$), RKO found the target solution for 80\% of instances within a factor of two from the best performance, compared to 70\% for RKO-BRKGA and less than 50\% for the other methods.

\subsection{Node Capacitated Graph Partitioning Problem} \label{sec:ncgpp}

The node-capacitated graph partitioning problem (NCGPP), also known as the handover minimization problem, can be described as follows. Let $B$ denote the set of base stations, where $T_b$ represents the total traffic handled by base station $b \in B$ and its connected transceivers. Additionally, let $N$ be the set of Radio Network Controllers (RNCs), where each RNC, denoted by $r \in N$, has a maximum traffic capacity of $C_r$. Furthermore, define $H_{b_1, b_2}$ as the total number of handovers between base stations $b_1$ and $b_2$ ($b_1, b_2 \in B, b_1 \neq b_2$). Note that $H_{b_1, b_2}$ and $H_{b_2, b_1}$ may differ.

The objective of the handover minimization problem is to assign each base station $b \in B$ to a specific RNC $r \in N$, such that the total number of handovers between base stations assigned to different RNCs is minimized. Let $\rho_b$ denote the index of the RNC to which base station $b$ is assigned, and $\Psi_r$ denote the indices of base stations assigned to RNC $r$. The assignments must satisfy the capacity constraints of each RNC, meaning that for all $r \in N$:

\[
\sum_{b \in \Psi_r} T_b \leq C_r.
\]

Mathematically, the problem can be formulated as finding the assignment of base stations to RNCs that minimize the total number of handovers between base stations assigned to different RNCs:

\[
\sum_{b_1, b_2 \in B \mid \rho_{b_1} \neq \rho_{b_2}} H_{b_1, b_2}.
\]

This optimization problem aims to balance the traffic load across the RNCs while minimizing the overall handover between base stations, which is crucial for efficient resource utilization and seamless user experiences in cellular networks.

One of the earliest contributions to the NCGPP came from \cite{FMSWW1998}, which introduced a branch-and-cut algorithm incorporating strong valid inequalities. Their approach was tested on three diverse applications: compiler design, finite element mesh computations, and electronic circuit layout. \cite{MEHROTRA19981} developed a branch-and-price algorithm for the NCGPP. Their experimental evaluation encompassed instances with 30 to 61 nodes and 47 to 187 edges. Notably, they solved these instances optimally, demonstrating the algorithm's effectiveness. A significant advancement in heuristic approaches came from \cite{DB2011}, which introduced a reactive GRASP coupled with path-relinking. This metaheuristic was tested on instances ranging from 30 to 82 nodes and 65 to 540 edges. The method performed well, matching CPLEX's optimal solutions in most cases while significantly reducing computational time. The proposed heuristic outperformed the CPLEX solver for larger instances, finding superior solutions. Finally, \cite{moran2013randomized} proposed three randomized heuristics. First, the authors use the GRASP with path-relinking proposed by \cite{mateus2011grasp} for the generalized quadratic assignment (GQAP). The NCGPP is a particular case of the GQAP, where facilities are the base stations, and locations are RNCs. They also developed a GRASP with evolutionary path-relinking (GevPR) and a BRKGA to solve the NCGPP. A benchmark set of 83 synthetic instances that mimic problems encountered in practice is proposed, varying in size from 20 to 400 nodes and 5 to 50 edges. The experiments show that the GevPR performed better than the other methods.

In this paper, we encode solutions of the NCGPP as a vector of $n=|B|+1$ random keys. The first $|B|$ positions represent each base station, while the last random key indicates the number of base stations assigned initially. Once sorted, the $|B|$ random keys will dictate the order in which base stations are assigned to the RNCs.

To decode a random-key vector and produce an assignment of base stations to RNCs, the following steps are computed: sort the random-key vector and then group base stations into RNCs, considering capacity constraints and costs. The assignment process starts by ensuring the first $\#N = \left \lceil  x_n \cdot |N| \right \rceil$ base stations are assigned to separate RNCs. For each subsequent base station ($|B| - \#N$), it searches for the best RNC that can accommodate the base station without exceeding capacity, evaluating the insertion cost based on the sum of handovers between the current base station and base stations that are already in the RNCs. If a suitable RNC is found, the base station is added, and the RNC’s capacity is updated. A penalty is added to the solution if no RNC can accommodate the base station. Finally, the decoder calculates the objective function value by summing the handover between base stations of different RNCs and the penalties for ungrouped base stations.

\begin{figure}[hbtp]
    \centering
    \includegraphics[width=0.75\linewidth]{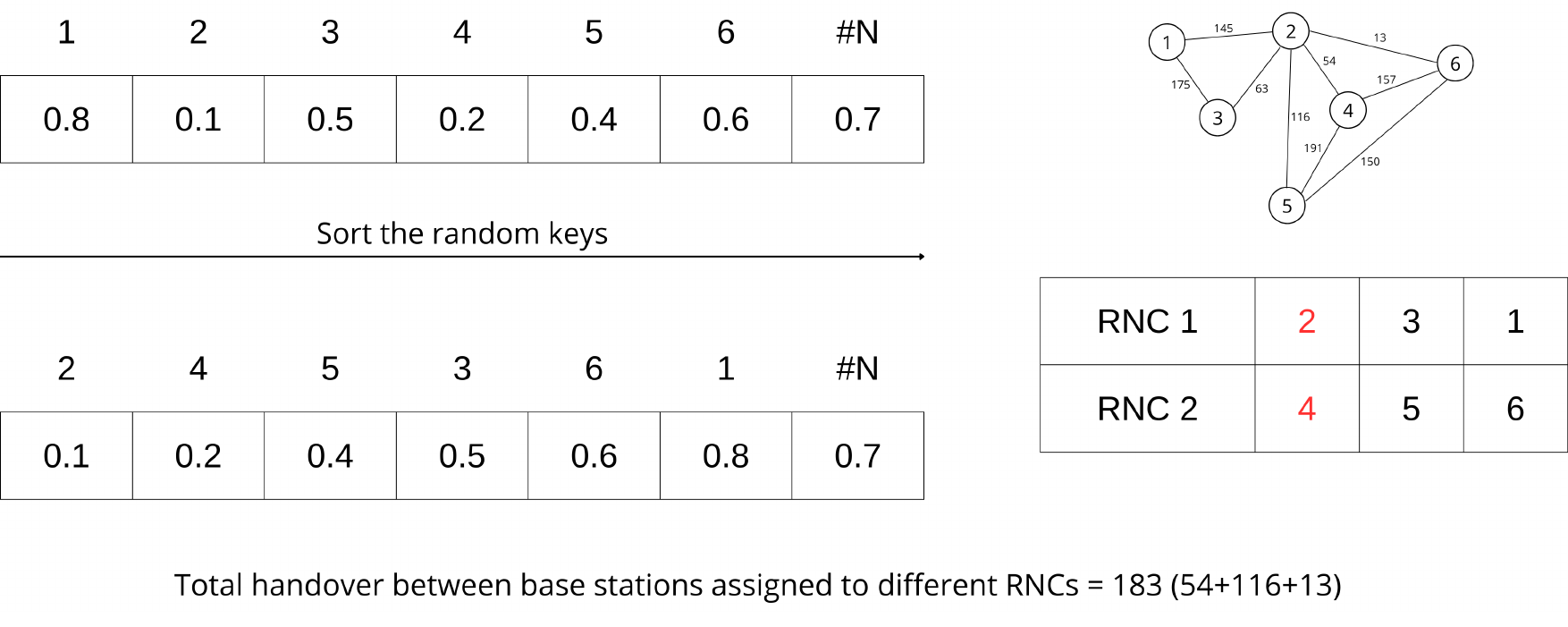}
    \caption{Example of the NCGPP decoder with six base stations and two RNCs.}
    \label{fig:decoderNCGPP}
\end{figure}

Figure \ref{fig:decoderNCGPP} shows an example of the NCGPP decoder with six base stations ($|B| = 6$) and two RNCs ($|N|=2$). In this example, $\#N = \left \lceil  0.7 \cdot 2 \right \rceil = 2$ and the ordered random-key vector provides the sequence of base stations: $2, 4, 5, 3, 6, 1$. From these $\#N$ and sequence, base stations $2$ and $4$ are allocated to separate RNCs. Base station $5$ is then allocated to RNC 2, as base station $5$ has 191 handovers with base station $4$ and only 116 with base station $2$. Base station $3$ is allocated to RNC 1, as it has no handover with base stations $4$ and $5$, only with base station $2$. Base station $6$ is allocated to RNC 2, which has 307 handovers (157 with base station $4$ and 150 with base station $5$) against 13 for RNC 1. Finally, base station $1$ is allocated to RNC 1 because it has only handovers with base stations $2$ and $3$.

We evaluated the performance of our proposed solution approach using the benchmark set with 83 instances proposed by \cite{moran2013randomized}, which varied in size from 20 to 400 base stations and 5 to 50 RNCs. First, we evaluated the performance of the commercial solver Gurobi (version 11.0) on the mathematical model presented in \cite{moran2013randomized}. Gurobi provided a baseline for comparison against our RKO methods. Next, we compared the results obtained by our RKO methods against methods from the literature, including GRASP for the GQAP \citep{mateus2011grasp}, GRASP with evolutionary path-relinking (GevPR) and BRKGA \citep{moran2013randomized}. 
We implemented a stopping criterion for the RKO methods based on a computational time limit equal to the number of base stations in each instance (measured in seconds). In contrast, we limited the execution time of the Gurobi solver to 1800 seconds per instance. Regarding the three heuristic methods from the literature, the authors allocated one hour of runtime for small instances and 24 hours for large instances using a Core 2 Duo processor with 2.2 GHz and 2.0 GB RAM.

Table \ref{tab:parameterHMP} presents the parameters of each metaheuristic used by the RKO to solve the NCGPP.

\begin{table}[ht]
\caption{Parameters of the RKO metaheuristics to solve the NCGPP.}
\label{tab:parameterHMP}
\centering
\tiny
\begin{tabularx}{\textwidth}{lXrrrrrrrr} \toprule
Parameters & Definition & BRKGA & GA & SA & ILS & VNS & GRASP & PSO & LNS \\
\midrule
$p$ & population size & 1597 & 1000 &  &  &  &  & 50 &  \\
$p_e$ & elite set & 0.10 &  &  &  &  &  &  &  \\
$p_m$ & mutant set & 0.20 &  &  &  &  &  &  &  \\
$\rho$ & inherit probability & 0.70 &  &  &  &  &  &  &  \\
$p_c$ & crossover probability &  & 0.85 &  &  &  &  &  &  \\
$\mu$ & mutation probability &  & 0.002 &  &  &  &  &  &  \\
$T_0$ & initial temperature &  &  & 1000000 &  &  &  &  & 1000 \\
$\mathit{SA}_{\mathit{max}}$ & number of iterations &  &  & 1000 &  &  &  &  &  \\
$\alpha$ & cooling rate &  &  & 0.99 &  &  &  &  & 0.90 \\
$\beta_{min}$ & minimum rate of shaking &  &  & 0.005 & 0.005 & 0.005 &  &  & 0.10 \\
$\beta_{max}$ & maximum rate of shaking &  &  & 0.05 & 0.10 &  &  &  & 0.30 \\
$k_{max}$ & number of neighborhoods &  &  &  &  & 10 &  &  &  \\
$h_s$ & start grid dense &  &  &  &  &  & 0.125 &  &  \\
$h_e$ & end grid dense &  &  &  &  &  & 0.00012 &  &  \\
$c_1$ & cognitive coefficient &  &  &  &  &  &  & 2.05 &  \\
$c_2$ & social coefficient &  &  &  &  &  &  & 2.05 &  \\
$w$ & inertia weight &  &  &  &  &  &  & 0.73 & 
\\ \bottomrule
\end{tabularx}
\end{table}

Table \ref{tab:resultsHMP} summarises the results for the NCGPP. The Gurobi solver successfully identified optimal solutions for all smaller instances (up to 40 nodes) and proved optimality for three instances with 100 nodes. For larger instances, the average optimality gap was 15.24\% (100-node instances), 80\% (200-node instances), and 99\% (400-node instances). Gurobi obtained the best-known solution ($\mathit{BKS}$) in 49 instances.

The GevPR \citep{moran2013randomized} heuristic and RKO methods, except RKO-BRKGA, found optimal solutions for the smallest instances. RKO-BRKGA failed to identify the optimum in one instance. Across the benchmark set, GevPR identified the $\mathit{BKS}$ in 56\% of instances (47 out of 83). The individually run RKO metaheuristics generally outperformed GevPR, with RKO-PSO being the exception, finding the $\mathit{BKS}$ in 42 instances. RKO-SA demonstrated the best performance, identifying the $\mathit{BKS}$ in 86\% of instances, with a best relative percentage deviation ($\mathit{RPD}_{\mathit{best}}$) of 0.02\% and an average computational time of 73 seconds. The parallel implementation of RKO further enhanced the performance of the individual metaheuristics, identifying the $\mathit{BKS}$ in 76 instances (91\%) within 70 seconds. The average $\mathit{RPD}_{\mathit{best}}$ was 0.01\% and the average $\mathit{RPD}_{\mathit{avg}}$ was 0.08\%.

\begin{table}[ht]
\centering
\caption{Summary of the NCGPP results for the benchmark instances. }
\label{tab:resultsHMP}
\begin{tabular}{lrrrrr} \toprule
Method               & \textit{Best}     & $\mathit{RPD}_{\mathit{best}}$ & $\mathit{RPD}_{\mathit{avg}}$  & best found at (s) & $\#\mathit{BKS}$ \\ \midrule
Gurobi       & 232534.93 & 24.39    & -     & 487.83            & 49    \\
GRASP$^1$         & 157464.01 & 4.46     & -     & -          & 47    \\
GevPR$^2$    & 139098.19 & 0.64     & -     & -          & 47    \\
BRKGA$^2$        & 142737.82 & 1.69     & -     & -          & 37    \\
RKO          & 136605.66 & 0.01     & 0.08  & 70.71             & \textbf{76}    \\
RKO-BRKGA    & 137395.28 & 0.20     & 11.11 & 75.91             & 56    \\
RKO-SA       & 136680.94 & 0.02     & 0.12  & 73.04             & 72    \\
RKO-GRASP    & 137690.41 & 0.21     & 0.55  & 91.52             & 59    \\
RKO-ILS      & 137994.82 & 0.29     & 0.56  & 80.59             & 55    \\
RKO-VNS      & 137701.45 & 0.21     & 0.56  & 78.79             & 58    \\
RKO-PSO      & 140043.08 & 0.91     & 1.60  & 27.22             & 42    \\
RKO-GA       & 137656.60 & 0.27     & 2.18  & 66.81             & 50    \\ 
RKO-LNS	    & 137827.69	& 0.26      & 0.50	& 82.79	     &58 \\ \bottomrule
\multicolumn{6}{l}{\cite{mateus2011grasp}$^1$ \cite{moran2013randomized}$^2$ }
\end{tabular}
\end{table}

\begin{figure}[ht]
    \centering
    \includegraphics[width=1.0\linewidth]{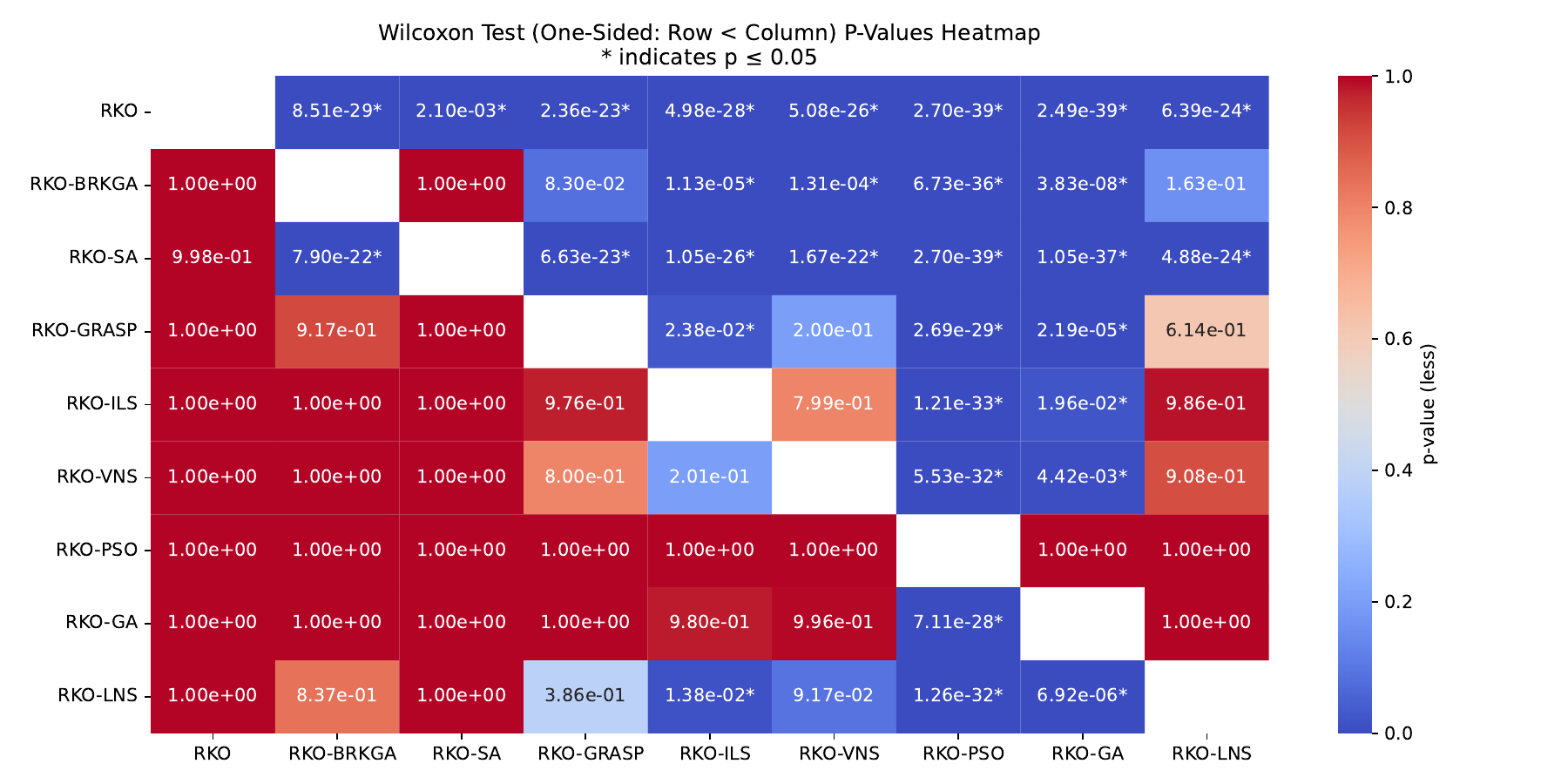}
    \caption{Heat map of the results ($p$-values) of the Wilcoxon signed-rank test for the RKO methods to solve the NCGPP.}
    \label{fig:wilcoxonHMP}
\end{figure}

As in \Cref{sec:aNpMP}, we subjected the RKO methods to statistical evaluation using the Wilcoxon signed-rank test. Figure \ref{fig:wilcoxonHMP} shows the results of all pair comparisons, where cells highlighted in blue with an asterisk (*) denote cases with statistically significant differences.
Our analysis indicated that the parallel RKO algorithm exhibited significant differences from all individually applied metaheuristics. Additionally, we observed no statistically significant differences among RKO-BRKGA, RKO-GRASP, RKO-ILS, RKO-VNS, and RKO-LNS metaheuristics.

\begin{figure}[ht]
    \centering
    \includegraphics[width=0.80\linewidth]{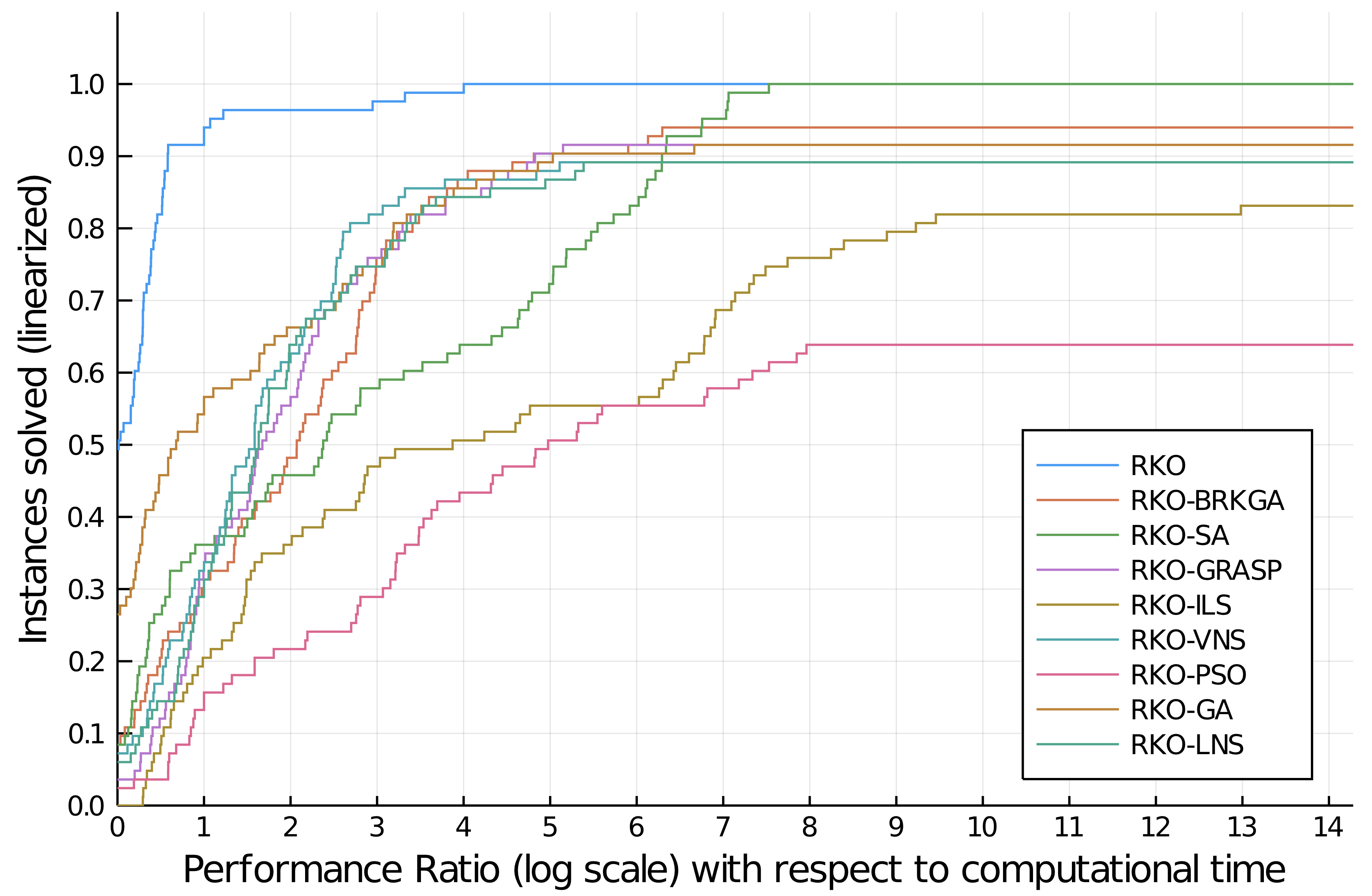}
    \caption{Performance profile of runtime for RKO methods to solve the NCGPP.}
    \label{fig:perfprofHMP}
\end{figure}

We also evaluated the computational time performance of the RKO methods with the performance profile plot. Figure \ref{fig:perfprofHMP} presents the performance profiles of the RKO methods on a $\log_2$ scale. We utilized a performance profile incorporating a convergence test to assess the methods' accuracy (tolerance threshold was $\mathit{RPD}_{\mathit{best}} = 1\%$). The RKO method showed superior computational efficiency and solution quality, outperforming individual metaheuristics when applied to benchmark instances. For a 1\% accuracy threshold, RKO found the target solution across all instances with a performance ratio of $\tau = 16$ (equivalent to $\log_2(\tau)=4$). In comparison, the RKO-SA, the only metaheuristic that also found all the target solutions, has a performance ratio of $\tau = 181$ ($\log_2(\tau) = 7.5$). Furthermore, RKO emerged as the most efficient solver for 50\% of the instances, while RKO-BRKGA showed in 28\% of cases, and other methods each accounted for less than 10\%. At a performance ratio of $\tau = 2$ ($\log_2(\tau) = 1$), RKO identified the target solution for 92\% of instances within twice the time of the best performer, compared to 55\% for RKO-BRKGA and under 40\% for the remaining methods.

\subsection{Tree Hub Location Problem} 
\label{sec:thlp}

The Tree Hub Location Problem (THLP) is an optimization problem that defines a set of $p_h$ hub nodes in a network, which are then connected by an undirected tree  \citep{contreras2010}. Each non-hub node must be assigned to one of the designated hubs, and all flows between nodes must travel through the hub network. Each arc in the network has an associated transportation cost per unit of flow, and there is a known demand between each pair of nodes in the network.

The objective function of the THLP has two main components. The first component calculates the transportation costs incurred by routing all demands from non-hub nodes to their assigned hubs. The second component accounts for the costs associated with the flows within the hub tree. A discount factor is applied to the latter component to reflect the potential economies of scale or efficiency gains inherent in the transportation model.

The literature on THLP has explored a variety of solution approaches, both exact and heuristic. Exact methods have included Mixed Integer Linear Programming (MILP) models \citep{contreras2009tight, contreras2010} and Benders decomposition techniques \citep{de2013improved}. On the heuristic side, researchers have proposed approaches such as the primal heuristic \citep{contreras2009tight} and the Biased Random-Key Genetic Algorithm (BRKGA) \citep{pessoa2017biased}. Furthermore, variants of the THLP have been studied in the literature, expanding the problem scope \citep{de2015exact,kayicsouglu2021multiple}. Notably, the BRKGA has emerged as a state-of-the-art method for the THLP, finding the best-known results in the literature.

\begin{figure}[htbp]
    \centering
    \includegraphics[width=0.75\linewidth]{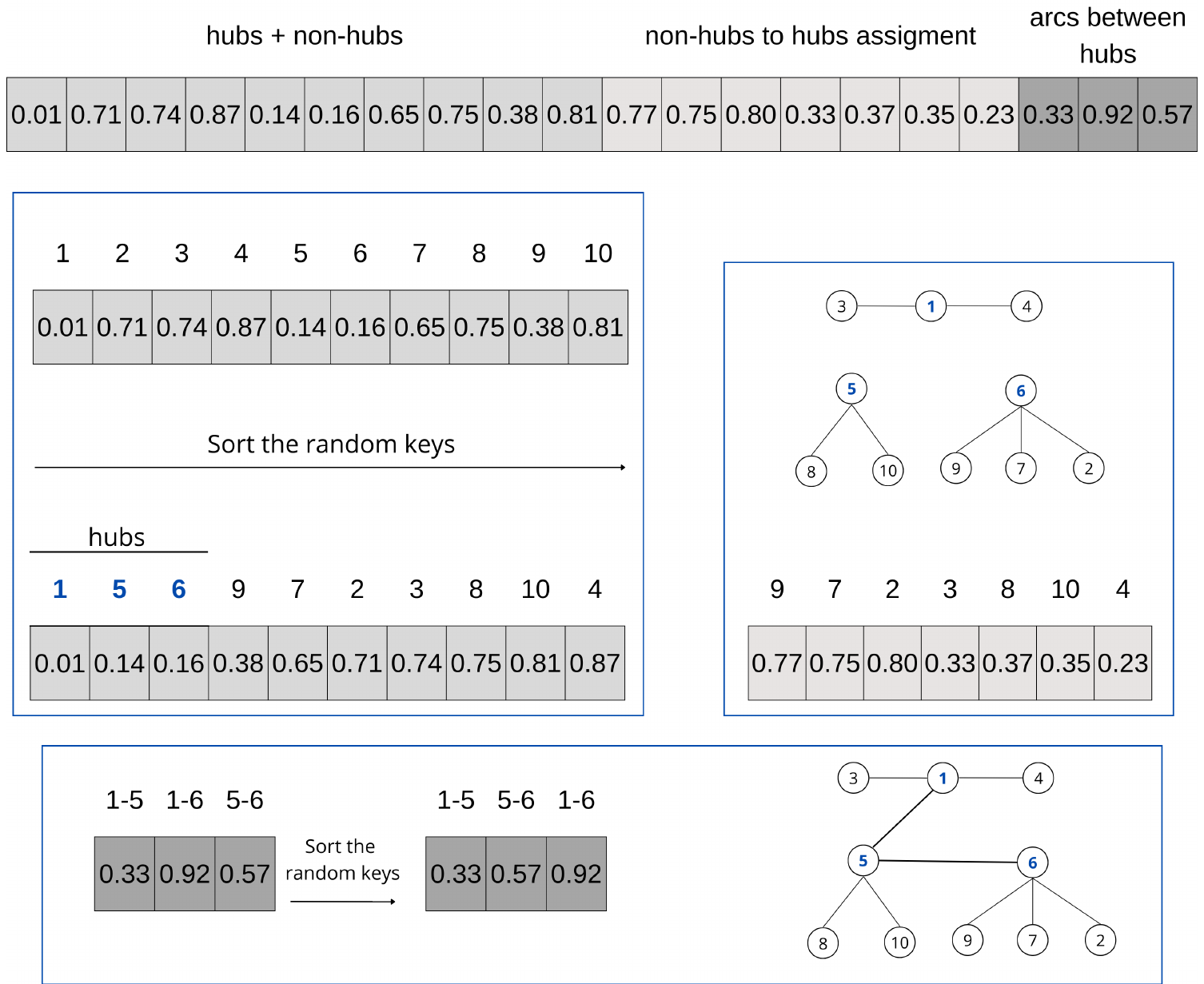}
    \caption{Example of the THLP decoder with ten nodes and three hubs.}
    \label{fig:decoderTHLP}
\end{figure}

In this study, we adopt the encoding proposed by \citet{pessoa2017biased} for our analysis. Consequently, the THLP solution is denoted by a random-key vector composed of three distinct segments. As depicted in Figure \ref{fig:decoderTHLP}, the first segment of the vector, of dimension $|N|$, corresponds to the network nodes. The second segment, sized at $|N| - p$, assigns non-hub nodes to their corresponding hubs. Finally, the third segment, sized at $p(p-1)/2$, represents all possible pairs of edges connecting the hubs.

Our decoder is also based on the decoder presented in \citet{pessoa2017biased}. It begins by sorting the vector’s first segment to delineate hubs and non-hubs: the initial $p$ positions denote hubs, while the subsequent $|N|-p$ positions are the non-hub nodes. Subsequently, the second segment undertakes the task of assigning non-hub nodes to their respective hubs. Achieving this involves partitioning the interval $[0, 1]$ into $p$ equal intervals, each associated with a specific hub per the preceding ordering. Lastly, the decoder arranges the random keys corresponding to inter-hub arcs in ascending order. The resulting tree structure is then constructed using the Kruskal algorithm applied to the sorted arcs. Figure \ref{fig:decoderTHLP} shows an example of the THLP decoder with $|N|=10$ and $p=3$.

To evaluate the efficacy of our proposed RKO, we utilized two sets of benchmark instances, referred to AP (Australian Post) and CAB (Civil Aeronautics Board), as described in previous studies \citep{contreras2009tight, contreras2010}. The AP dataset contains information on postal districts across Australia, while the CAB dataset includes information on the American cities with the highest volume of airline passenger traffic. These datasets have 126 instances divided between smaller-scale instances (10, 20, and 25 nodes), medium (40, 50, and 60 nodes), and large-scale (70, 75, 90, and 100 nodes). Each instance has three different values of $p_h$ (3, 5, and 8) and discount factors (0.2, 0.5, and 0.8). We compared our RKO methods and methods from the literature, such as an exact method \citep{contreras2009tight,contreras2010}, a primal heuristic \citep{contreras2009tight}, and a BRKGA \citep{pessoa2017biased}. We implemented a stop condition based on a maximum computational time for our RKO methods. This time limit was set to correspond with the number of network nodes in each instance, measured in seconds.

Table \ref{tab:parameterTHLP} presents the parameters of each metaheuristic used by the RKO to solve the THLP.

\begin{table}[ht]
\caption{Parameters of the RKO metaheuristics to solve the THLP.}
\label{tab:parameterTHLP}
\centering
\tiny
\begin{tabularx}{\textwidth}{lXrrrrrrrr} \toprule
Parameters & Definition & BRKGA & GA & SA & ILS & VNS & GRASP & PSO & LNS \\
\midrule
$p$ & population size & 1597 & 600 &  &  &  &  & 200 &  \\
$p_e$ & elite set & 0.15 &  &  &  &  &  &  &  \\
$p_m$ & mutant set & 0.20 &  &  &  &  &  &  &  \\
$\rho$ & inherit probability & 0.70 &  &  &  &  &  &  &  \\
$p_c$ & crossover probability &  & 0.99 &  &  &  &  &  &  \\
$\mu$ & mutation probability &  & 0.005 &  &  &  &  &  &  \\
$T_0$ & initial temperature &  &  & 1000000 &  &  &  &  & 1000000 \\
$\mathit{SA}_{\mathit{max}}$ & number of iterations &  &  & 1500 &  &  &  &  &  \\
$\alpha$ & cooling rate &  &  & 0.99 &  &  &  &  & 0.97 \\
$\beta_{min}$ & minimum rate of shaking &  &  & 0.01 & 0.05 & 0.005 &  &  & 0.10 \\
$\beta_{max}$ & maximum rate of shaking &  &  & 0.05 & 0.20 &  &  &  & 0.30 \\
$k_{max}$ & number of neigborhoods &  &  &  &  & 10 &  &  &  \\
$h_s$ & start grid dense &  &  &  &  &  & 0.125 &  &  \\
$h_e$ & end grid dense &  &  &  &  &  & 0.00012 &  &  \\
$c_1$ & cognitive coefficient &  &  &  &  &  &  & 2.05 &  \\
$c_2$ & social coefficient &  &  &  &  &  &  & 2.05 &  \\
$w$ & inertia weight &  &  &  &  &  &  & 0.73 & 
\\ \bottomrule
\end{tabularx}
\end{table}

Tables \ref{tab:THLPsmall} and \ref{tab:THLPlarge} present the results for the THLP. The exact methods proposed by \cite{contreras2009tight, contreras2010} proved optimal solutions for 59 out of 63 small instances. The RKO algorithm identified these optimal solutions and discovered the best-known solutions ($\mathit{BKS}$) for the remaining four instances. All RKO metaheuristics performed well for small instances, with the exception of BRKGA, which found only 37 $\mathit{BKS}$. Similarly, the BRKGA proposed by \cite{pessoa2017biased} found only 45 $\mathit{BKS}$. Consequently, the methodologies of the other metaheuristics demonstrated greater efficacy than BRKGA for this set of THLP instances. RKO methods also obtained the $\mathit{BKS}$ for medium and large instances. While the primal heuristic \citep{contreras2009tight} identified the $\mathit{BKS}$ for only 29 out of 63 instances, the parallel RKO algorithm successfully found the $\mathit{BKS}$ for 60 instances. RKO-LNS found the $\mathit{BKS}$ for the remaining three instances. Besides, RKO-GRASP, RKO-ILS, and RKO-VNS each found the $\mathit{BKS}$ in one of these instances. Notably, RKO was the method that obtained the lowest value for $RPD_{best}$ and $RPD_{aver}$.

\begin{table}[ht]
\centering
\caption{Summary of the THLP results for the CAP and AP small instances.}
\label{tab:THLPsmall}
\begin{tabular}{lrrrrr} \toprule
Method               & \textit{Best}     & $\mathit{RPD_{best}}$ & $\mathit{RPD_{avg}}$  & best found at (s) & $\#\mathit{BKS}$ \\ \midrule
Exact method$^{1}$ & 26620.58 & 0.04 & - & 1436.51 & 59 \\
BRKGA$^2$ & 26685.72 & 0.22 & - & 53.93 & 45 \\
RKO & 26608.76 & 0.00 & 0.04 & 1.04 & \textbf{63} \\
RKO-BRKGA & 26826.06 & 0.66 & 1.59 & 4.19 & 37 \\
RKO-SA & 26609.03 & 0.00 & 0.05 & 3.54 & 62 \\
RKO-GRASP & 26608.76 & 0.00 & 0.04 & 1.69 & \textbf{63} \\
RKO-ILS & 26608.76 & 0.00 & 0.07 & 1.29 & \textbf{63} \\
RKO-VNS & 26608.76 & 0.00 & 0.00 & 1.08 & \textbf{63} \\
RKO-PSO & 26608.76 & 0.00 & 0.00 & 1.69 & \textbf{63} \\
RKO-GA & 26612.17 & 0.02 & 0.09 & 3.64 & 58 \\
RKO-LNS & 26608.76 & 0.00 & 0.00 & 1.30 & \textbf{63}
 \\ \bottomrule
\multicolumn{6}{l}{\cite{contreras2009tight,contreras2010}$^1$ \cite{pessoa2017biased}$^2$ }
\end{tabular}
\end{table}

\begin{table}[ht]
\centering
\caption{Summary of the THLP results for the AP medium and large instances.}
\label{tab:THLPlarge}
\begin{tabular}{lrrrrr} \toprule
Method               & \textit{Best}     & $\mathit{RPD_{best}}$ & $\mathit{RPD_{avg}}$  & best found at (s) & $\#\mathit{BKS}$ \\ \midrule
Primal heuristic$^1$ & 66203.51 & 0.89 & - & 1616.26 & 29 \\
RKO & 65639.39 & 0.005 & 0.12 & 16.75 & \textbf{60} \\
RKO-BRKGA & 67281.20 & 2.71 & 5.10 & 9.62 & 1 \\
RKO-SA & 65940.80 & 0.46 & 1.02 & 30.04 & 35 \\
RKO-GRASP & 65651.48 & 0.03 & 0.31 & 28.64 & 56 \\
RKO-ILS & 65642.42 & 0.009 & 0.57 & 21.53 & 58 \\
RKO-VNS & 65642.51 & 0.009 & 0.12 & 27.26 & 59 \\
RKO-PSO & 65683.33 & 0.07 & 0.20 & 24.90 & 48 \\
RKO-GA & 65777.66 & 0.22 & 0.71 & 30.06 & 34 \\
RKO-LNS & 65642.73 & 0.01 & 0.10 & 23.66 & 59
\\ \bottomrule
\multicolumn{6}{l}{\cite{contreras2009tight}$^1$ }
\end{tabular}
\end{table}

\begin{figure}[htbp]
    \centering
    \includegraphics[width=1.0\linewidth]{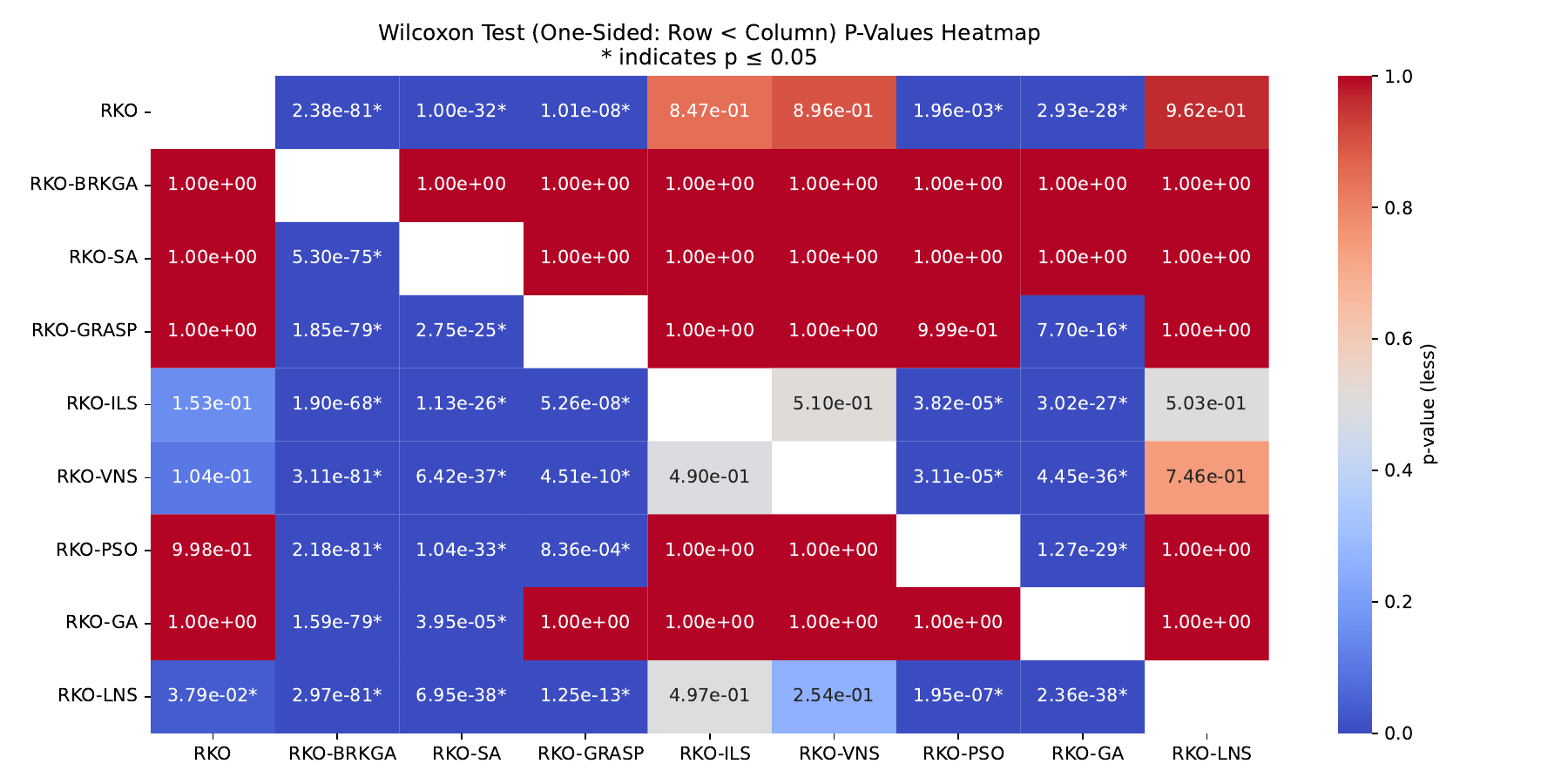}
    \caption{Heat map of the results ($p$-values) of the Wilcoxon signed-rank test for the RKO methods to solve the THLP.}
    \label{fig:wilcoxonTHLP}
\end{figure}

The Wilcoxon signed-rank test corroborates this analysis. The results, shown in Figure \ref{fig:wilcoxonTHLP}, indicate that RKO is statistically significantly different from RKO-BRKGA, RKO-SA, RKO-GRASP, RKO-PSO, and RKO-GA. However, no statistically significant difference was found between RKO and RKO-ILS or between RKO and RKO-VNS. We also found a statistically significant difference between RKO-LNS and RKO, with the individual metaheuristic achieving better performance ranks than the RKO.

\begin{figure}[htbp]
    \centering
    \includegraphics[width=0.78\linewidth]{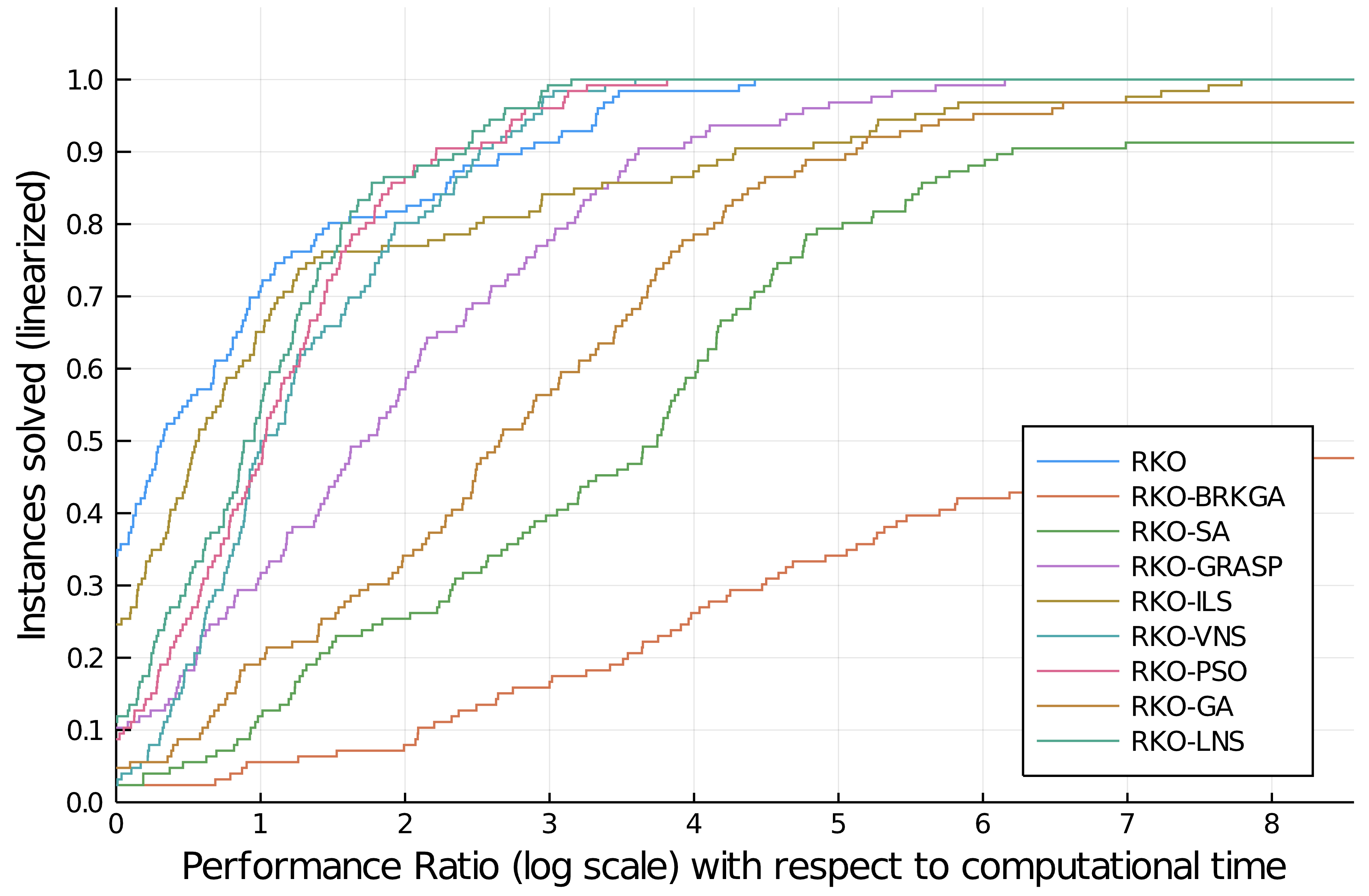}
    \caption{Performance profile of runtime for RKO methods to solve the THLP.}
    \label{fig:perfprofTHLP}
\end{figure}

Finally, we evaluated the computational time performance of the RKO methods with the performance profile. Figure \ref{fig:perfprofTHLP} presents the performance profiles of the RKO methods on a $\log_2$ scale. We employed a performance profile that included a convergence test to evaluate the accuracy of the methods, with a tolerance threshold set at \(\mathit{RPD}_{\mathit{best}} = 1\%\).
The RKO method consistently outperformed the other methods individually. However, its superiority is less pronounced compared to the previous two problems. RKO was the most efficient method for solving 35\% of the instances, while RKO-ILS was the most efficient in 25\% of the cases. With a performance factor of \(\tau = 2\) (\(\log_2(\tau) = 1\)), RKO solved up to 70\% of the instances, and RKO-ILS solved 65\%. RKO-VNS, RKO-PSO, and RKO-LNS each solved 50\% of the instances. However, when \(\tau = 4\) (\(\log_2(\tau) = 2\)), RKO-LNS and RKO-PSO solved more instances than RKO (86\% versus 82\%). These methods and RKO-VNS achieved 100\% of solved instances with a better performance factor than RKO.

\subsection{RKO with Q-Learning}

This section presents the results of applying RKO with online parameter control using the Q-Learning method. The sets of possible values for each metaheuristic parameter were derived from Tables \ref{tab:parameterAlphaNpMP}, \ref{tab:parameterHMP}, and \ref{tab:parameterTHLP}. These parameter sets were then used to create a Markov Decision Process (MDP) for each metaheuristic, enabling Q-Learning to identify the most appropriate configurations (policy) for each problem, instance, and at various stages of the search process.

In the RKO framework with Q-Learning, users only need to implement problem-specific decoders, while the proposed approach dynamically adjusts the metaheuristic parameters during the search. This adaptive behavior allows the exploration of the random-key solution space, improving the likelihood of finding high-quality solutions.

Table \ref{tab:RKO-QL} presents the results of RKO using both parameter tuning and parameter control. The performance of RKO with Q-Learning was similar to that of the offline parameter configuration version. Both approaches often effectively identified the best solutions across the three optimization problems, exhibiting comparable average best solutions. 
However, the relative percentage deviations of the RKO with parameter tuning are better. 
The RKO with Q-Learning also exhibited longer computational times in the THLP and NCGPP, which can be attributed to the need for multiple iterations to learn an effective policy for parameter adjustment. Wilcoxon’s statistical test indicated a significant difference between the RKO versions ($p$-value $< 0.05$).

\begin{table}[htbp]
\caption{Comparison of RKO results with parameter tuning and parameter control.}
\label{tab:RKO-QL}
\centering
\begin{tabularx}{\textwidth}{llrrrrrr} \toprule
 & Method & \textit{Best} & $\mathit{RPD_{best}}$ & $\mathit{RPD_{avg}}$ & best found at (s) & $\#\mathit{BKS}$ & $p$-value \\ \midrule
 
\multirow{2}{*}{THLP} & RKO & 46124.08 & 0.003 & 0.080 & 8.89 & 123 & \multirow{2}{*}{$2.05\,\mathrm{e}{-05}$} \\
 & RKO-QL & 46131.51 & 0.019 & 0.136 & 12.16 & 117 &  \\
 &  &  &  &  &  &  &  \\
 
\multirow{2}{*}{$\alpha$NpMP} & RKO & 84300.39 & 0.003 & 0.018 & 28.70 & 65 & \multirow{2}{*}{$3.26\,\mathrm{e}{-04}$} \\
 & RKO-QL & 84314.90 & 0.015 & 0.056 & 26.38 & 55 &  \\
 &  &  &  &  &  &  &  \\
 
\multirow{2}{*}{NCGPP} & RKO & 136605.66 & 0.009 & 0.076 & 70.71 & 76 & \multirow{2}{*}{$3.43\,\mathrm{e}{-12}$} \\
 & RKO-QL & 137284.36 & 0.109 & 0.190 & 92.74 & 65 & \\ \bottomrule

\end{tabularx}
\end{table}

\section{Conclusion}\label{sec:conclusion}

This paper introduced the Random-Key Optimizer (RKO), a novel and versatile optimization framework designed to tackle a wide range of combinatorial optimization problems. By encoding solutions as vectors of random keys and utilizing problem-specific decoders, RKO provides a unified approach that integrates multiple metaheuristics, including simulated annealing, iterated local search, and greedy randomized adaptive search procedures. The RKO framework's modular design allows it to adapt to various optimization challenges, consistently yielding high-quality solutions.

Our extensive testing on NP-hard problems such as the $\alpha$-neighborhood $p$-median problem, the tree of hubs location problem, and the node-capacitated graph partitioning problem demonstrated RKO's effectiveness in producing optimal or near-optimal solutions efficiently. The framework’s flexibility in incorporating diverse metaheuristics, coupled with its superior performance across different problem domains, underscores its robustness and potential as a foundational tool for combinatorial optimization.

The innovation of combining random-key encoding with a modular metaheuristic approach opens new avenues for addressing complex optimization problems that were previously intractable. Tuning the parameters of metaheuristics can be a time-consuming and computationally expensive task. To address this, we enhance the framework by developing a hybrid approach that integrates RKO with machine learning techniques to predict optimal parameter configurations during the search process. Future research could focus on incorporating adaptive mechanisms that dynamically select the most effective metaheuristic based on the problem instance. Expanding the elite solution pool's role in guiding the search process and exploring alternative solution-sharing strategies could further improve the framework's performance.

As the field of optimization continues to evolve, the RKO framework is well-positioned to serve as a powerful tool for researchers and practitioners alike. Its adaptability and effectiveness suggest broad applicability in other combinatorial optimization challenges, including emerging fields such as network design, logistics, and bioinformatics. Future research could apply a linear programming method to explore the search space of random keys, potentially enhancing the framework's efficiency. By continuing to refine and expand this framework, we can unlock new possibilities for solving optimization problems with greater efficiency and effectiveness, ultimately advancing the state of the art in operations research.

\backmatter
\bmhead{Acknowledgements}

The authors thank the anonymous reviewers for their time and constructive comments. Their feedback was essential for improving the quality and clarity of this manuscript.


\section*{Declarations}

\noindent \textbf{Funding}: This study was funded by FAPESP under grants 2018/15417-8, 2022/05803-3, and 2024/08848-3, and Conselho Nacional de Desenvolvimento Cient\'ifico e Tecnol\'ogico (CNPq) under grants 312747/2021-7 and 405702/2021-3.

\vspace{0.2cm}

\noindent \textbf{Data availability}: The instances and the source code are available at: \url{https://github.com/RKO-solver}

\vspace{0.2cm}

\noindent \textbf{Conflict of Interest}: The authors declare that they have no conflict of interest.

\vspace{0.2cm}

\noindent \textbf{Statement of human rights}: This article does not contain any studies with human participants or animals performed by any of the authors.

\vspace{0.2cm}

\noindent \textbf{Informed consent} was obtained from all individual participants included in the study.

\bibliography{bibliography}

\end{document}